\documentclass[lettersize,journal]{IEEEtran}
\usepackage{amsmath,amsfonts}
\usepackage{algorithmic}
\usepackage{array}
\usepackage[caption=false,font=normalsize,labelfont=sf,textfont=sf]{subfig}
\usepackage{textcomp}
\usepackage{stfloats}
\usepackage{url} 
\usepackage{verbatim}
\usepackage{graphicx}
\def\BibTeX{{\rm B\kern-.05em{\sc i\kern-.025em b}\kern-.08em
    T\kern-.1667em\lower.7ex\hbox{E}\kern-.125emX}}
\usepackage{balance}
\usepackage{booktabs}
\usepackage{multirow}
\usepackage{amsthm,amsmath,amssymb}
\usepackage{mathrsfs}
\usepackage{threeparttable}
\usepackage{cite}
\usepackage{xcolor}
\usepackage{colortbl} 
\usepackage{hyperref}
\hypersetup{urlcolor=magenta, % url
 	citecolor=green, % citation
 	linkcolor=red, % table of contents, inner color
 	colorlinks=true,  }
\begin{document}
\title{Learning Remote Sensing Object Detection with Single Point Supervision}
\author{Shitian~He, Huanxin~Zou, Yingqian~Wang, Boyang~Li, Xu~Cao, Ning~Jing
\thanks{This work was supported by the National Natural Science Foundation of China under Grant 62071474. (Corresponding author: Huanxin Zou).
	
S.~He, H.~Zou, Y.~Wang, B.~Li, X.~Cao and N.~Jing are with the College of Electronic Science and Technology, National University of Defense Technology, Changsha, China. (E-mails: \{heshitian19, zouhuanxin, wangyingqian16, liboyang20, cx2020, ningjing\}@nudt.edu.cn
}}

%\markboth{SUBMITTED TO IEEE TRANSACTIONS ON GEOSCIENCE AND REMOTE SENSING}
%{How to Use the IEEEtran \LaTeX \ Templates}
%\markboth{Journal of \LaTeX\ Class Files,~Vol.~18, No.~9, April~2023}%

\maketitle

\begin{abstract}
Pointly Supervised Object Detection (PSOD) has attracted considerable interests due to its lower labeling cost as compared to box-level supervised object detection. However, the complex scenes, densely packed and dynamic-scale objects in Remote Sensing (RS) images hinder the development of PSOD methods in RS field. In this paper, we make the first attempt to achieve RS object detection with single point supervision, and propose a PSOD method tailored for RS images. Specifically, we design a point label upgrader (PLUG) to generate pseudo box labels from single point labels, and then use the pseudo boxes to supervise the optimization of existing detectors. Moreover, to handle the challenge of the densely packed objects in RS images, we propose a sparse feature guided semantic prediction module which can generate high-quality semantic maps by fully exploiting informative cues from sparse objects.
Extensive ablation studies on the DOTA dataset have validated the effectiveness of our method. Our method can achieve significantly better performance as compared to state-of-the-art image-level and point-level supervised detection methods, and reduce the performance gap between PSOD and box-level supervised object detection. Code is available at \url{https://github.com/heshitian/PLUG}.

\end{abstract}


\section{Introduction}
\label{sec:intro}

\IEEEPARstart{R}{emote} Sensing Object detection (RSOD) plays an important role in many fields, such as national defense and security, resource managing and emergency rescuing. With the development of deep learning, many deep-netural-network (DNN) based detection methods \cite{hou2022refined, cheng2018learning, huang2022general,kong2020foveabox, liu2022progressive, li2021gsdet, han2021align} were proposed and achieved promising performance. 
Besides, a number of Remote Sensing (RS) datasets (e.g., HRSC2016 \cite{HRSC}, NWPU VHR-10 \cite{NWPUVHR10} and DOTA series \cite{DOTA}) containing accurate and rich annotations were proposed to develop and benchmark RSOD methods. In these datasets, accurate location, scale, category and quantity information of objects are provided and greatly facilitate the development of RSOD. However, such rich annotation formats will lead to expensive labor costs when RSOD methods are transferred to the new RS data (e.g., images captured by new satellites).

To reduce the labor costs of annotating new RS data, researchers explored image-level annotations where only category information of objects are provided, and introduced image-level supervised detection methods \cite{bilen2016weakly, tang2017multiple, tang2018pcl, zeng2019wsod2, zhang2018adversarial, xie2021online}. These methods generally detect objects in a ``find-and-refine" pipeline, i.e., the coarse positions of objects are firstly found, and the proposals are then generated and refined. However, due to the complex RS scenes and the lack of location, scale, and quantity information, it is highly challenging to achieve good RSOD performance based on image-level annotation.
Recently, single point annotation \cite{papadopoulos2017training, ren2020ufo, chen2022p2b, yu2022object} has attracted much attention. Different from image-level annotations, point labels can simultaneously provide category, quantity and coarse position information. The introduction of additional location and quantity information simplifies the original ``find-and-refine" pipeline to the ``refine-only" one, and thus reduces the difficulties of pseudo box generation. Besides, the labor cost of single point annotations is only about one-eighteenth of box-level labels \cite{chen2022p2b}, and is negligibly higher than image-level ones. Therefore, single point annotations have large potential in the detection field.

\begin{figure}[t]
	\centering
	\includegraphics[width= 8.5 cm]{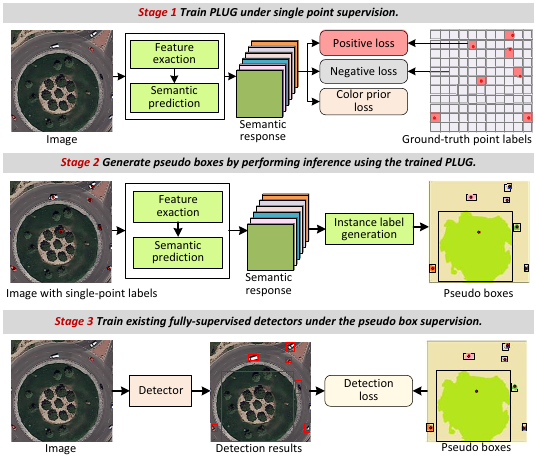}
	\caption{{An illustration of the training pipeline of our PSOD method. Firstly, our PLUG is trained under single point supervision. Then, pseudo box labels are generated by performing inference using the well-trained PLUG. Finally, existing fully supervised detectors (e.g., Faster-RCNN) are trained under the supervision of the generated pseudo boxes.}}
	\label{fig_first_0}
\end{figure}

Pointly supervised object detection (PSOD) is still in its infancy, with just a few methods \cite{papadopoulos2017training, ren2020ufo, chen2022p2b} being proposed in recent years. Papadopoulos et al. \cite{papadopoulos2017training} introduced center-click annotation, and used the error distribution between two clicks to estimate object scales. Ren et al. \cite{ren2020ufo} proposed a unified object detection framework that can handle different forms of supervision (e.g., tags, points, scribbles and boxes) simultaneously. Chen et al. \cite{chen2022p2b} predefined massive proposals in varied scales, aspect ratios and shaking degrees for each point label, and used multi-instance learning (MIL) to select and refine the most suitable proposals as the final results. 

A straightforward way to achieve pointly supervised RSOD is to directly apply existing PSOD methods to RS images. {These PSOD methods mainly follow the MIL pipeline, in which many proposals are preset for each point label, and then the optimal one is selected as the pseudo box label. However, this framework is unsuitable for the RSOD task due to the low recall of proposal bags caused by the extremely huge variation of scales and aspect ratios of RS objects. In this paper, we make the first attempt to achieve RSOD with single point supervision, and propose a point label upgrader (PLUG) to generate high quality pseudo box labels from single points. Specifically, the semantic response map is first learned under point-level supervision, and then pseudo boxes can be generated in shortest path paradigm. Due to the discard of proposal generation, our PLUG is less susceptible to the interference from varied scales and aspect ratios. Moreover, the dense and cluttered objects in RS images hamper the extraction of discriminative features, and thus degrade the qualities of generated pseudo boxes. Considering this issue, we propose a sparse feature guided semantic prediction (SemPred) module to extract general representations of sparse objects and utilize them to improve the quality of the pseudo boxes of dense objects. In this way, our PLUG can obtain more discriminative feature representations and improve the downstream detection performance.}

{By utilizing PLUG to transform single point labels into box-level ones, we can develop a PLUG-Det method to achieve PSOD tailored for RS images. The training pipeline of our PLUG-Det consists of three stages (as shown in Fig.~\ref{fig_first_0}). Firstly, our PLUG is trained under the single point supervision. Then, pseudo boxes are generated by performing inference using the well-trained PLUG. Finally, existing fully supervised detectors (e.g., Faster-RCNN) are trained using the pseudo boxes to achieve PSOD.}

In summary, our main contributions are as follows. 
\begin{itemize}
    \item We present the first study on single pointly supervised RSOD, and propose a simple yet effective method called PLUG to generate pseudo box annotations from single point ones.
    \item To handle the challenge of dense and clustered objects in RS images, we propose a sparse feature guided semantic prediction approach to enhance the discriminative feature representation capability of our PLUG.
    \item By using the generated pseudo boxes to train existing detectors (Faster-RCNN \cite{FasterRCNN} in this paper), our method (i.e., PLUG-Det) achieves promising detection performance, and outperforms many existing weakly supervised detectors.
%    \item Our PLUG-Dert framework can be generalized to different detectors, and can achieve 
\end{itemize}

The remainder of this paper is organized as follows. In Section \ref{sec:relatedworks}, we briefly review the related works. Section \ref{sec:methods} presents the details of the proposed method. Comprehensive experimental results are provided in Section \ref{sec:experiments}, and Section \ref{sec:conclusion} concludes this paper.

\section{Related Works}
\label{sec:relatedworks}

\subsection{Object Detection in Remote Sensing Images}
RSOD has been extensively investigated in the past decades. Since convolutional neural network (CNN) was proposed, deep learning based RSOD methods have achieved significant improvements \cite{li2020object}. Compared to objects in natural images, RS objects have some special characteristics \cite{yang2019scrdet}, including varied orientation, dynamic scales, densely packed arrangements, significant intra-class difference, etc. Therefore, RSOD methods generally focus on the solutions to the above unique issues.

Specifically, regarding the varied orientation issue, many researchers proposed new representation approaches, e.g., rotated bounding boxes \cite{ma2018arbitrary, jiang2018r, ding2019learning, xie2021oriented}, intersecting lines \cite{wei2020oriented, wei2020x}, key-points \cite{yi2021oriented, zhao2021polardet, dai2022ace} and rotated Gaussian distribution \cite{yang2021rethinking, yang2022kfiou}. Besides, some researchers proposed improved feature extraction modules \cite{han2021align, han2021redet, yang2021r3det}, novel loss functions \cite{yang2019scrdet, qian2021learning} and new angle regression mechanisms \cite{yang2020arbitrary, yang2021dense} to improve the detection performance on multi-oriented objects. 
Regarding the dynamic scales issue, Hu et al. \cite{hu2018relation} proposed a feature enhancement method that can extract more discriminative features containing structure, deep semantic and relation information simultaneously. In \cite{yang2019scrdet, deng2021extended}, multi-scale features were used to extract the scale-invariant representation of objects. Besides, Li et al. \cite{li2021gsdet} proposed a ground sample distance (GSD) identification sub-network and combined GSD information with the sizes of Regions of Interest (RoIs) to determine the physical size of objects.
Regarding the densely packed arrangement issue, Yang et al. \cite{yang2019clustered} proposed ClusDet, in which clustering regions were first extracted by a cluster proposal sub-network, and then fed to a specific detection sub-network for final prediction. Li et al. \cite{li2020density} proposed a density map guided detection method, where the density map can represent whether a region contains objects or not, and thus provide guidance for cropping images statistically.

Apart from the above studies, there are still many works trying to tackle other issues (e.g., excessive feature coupling \cite{fu2020point, liu2021dcl}, unbalanced label assignment \cite{xu2022rfla}, various aspect ratios \cite{zhu2020adaptive, liu2021center}) in RSOD.
Recently, Transformer-based object detection methods \cite{carion2020end, zhu2020deformable, li2022dn} have attracted much attention due to their strong modeling capability. Therefore, some Transformer-based RSOD methods \cite{dai2022ao2, wang2022advancing} have been proposed and achieved remarkable detection performance.  

The aforementioned methods improve the detection performance under box-level supervision. In this paper, we aim at relieving the labor cost of annotating RS images, and propose a single pointly supervised RSOD method.

\subsection{Image-level Supervised Object Detection}   
To relieve the burden of box-level labeling, numerous image-level supervised detection methods \cite{fasana2022weakly, cheng2020high, bilen2016weakly, tang2017multiple, tang2018pcl, zeng2019wsod2, zhang2018adversarial, xie2021online, feng2020progressive} were proposed, which can be categorized into class activation map (CAM) based and MIL-based methods. 

{CAM-based methods \cite{zhang2018adversarial, xie2021online} detect objects based on the class activation maps.
% \textcolor{black}{However, since CAM-based methods can only generate one proposal for each class \cite{shao2022deep},} it is not suitable for the RS images with multiple instances. 
Li et al. \cite{li2018deep} proposed a CAM-based detection framework, in which the mutual information between images was exploited, and the class-specific activation weights were learnt to better distinguish multi-class objects. Since CAM-based methods can only generate few proposals for each class \cite{shao2022deep}, it is not suitable for RS images with multiple instances.}

{MIL-based methods \cite{bilen2016weakly, tang2017multiple, tang2018pcl, zeng2019wsod2} generally utilize off-the-shelf proposal generators (e.g., selective search \cite{uijlings2013selective}, edge boxes \cite{zitnick2014edge} and sliding windows) to produce initial proposals, and then consider the proposal refinement process as an MIL problem to make final predictions \cite{shao2022deep}. For example, WSDDN \cite{bilen2016weakly} first generates proposals using edge boxes, then feeds the extracted features of proposals to two parallel branches for classification and detection scoring, respectively. The two obtained scores are used to classify positive proposals. Based on WSDDN, OICR \cite{tang2017multiple} uses selective search to generate proposals, and adds an {instance classification refinement} process to enhance the discriminatory capability of the instance classifier. PCL \cite{tang2018pcl} improves the original proposal bags to proposal clusters, so that spatially adjacent proposals with the same label can be assigned to the same category cluster. }
	
{In 2014, Zhang et al. \cite{zhang2014weakly} firstly transferred the image-level supervised detection methods into the RSOD field. Specifically, they first performed saliency-based segmentation and negative sample mining to generate initial training samples, and then proposed an iterative training approach to refine the samples and the detector gradually. On this basis, Han et al. \cite{han2014object} proposed a Bayesian framework to generate training samples, in which a deep Boltzmann machine was employed to extract the high-level features.
In image-level supervised RSOD field, the key challenging issues are the local discrimination, multi-instances and the imbalance between easy and difficult samples. Recent methods put efforts on the improvement against these issues. For example, regarding the local discrimination issue, Feng et al. \cite{feng2020tcanet} proposed a novel triple context-aware network, named TCANet, to learn complementary and discriminative visual features. Feng et al. \cite{feng2020progressive} subsequently proposed a progressive contextual instance refinement method. Qian et al. \cite{qian2023semantic} proposed a semantic segmentation guided pseudo label mining module to mine high-quality pseudo ground truth instances. Regarding the multi-instances issue, Wang et al. \cite{wang2021multiple} proposed a unique multiple instance graph learning framework. Feng et al. \cite{feng2022weakly} proposed to utilize the rotation-consistency to pursue all possible instances. Wang et al. \cite{wang2023mol} developed a novel multi-view noisy learning framework, named MOL, which uses reliable object discovery and progressive object mining to reduce the background interference and tackle the multi-instance issue. For the imbalanced easy and difficult samples, Yao et al. \cite{yao2020automatic} performed dynamic curriculum learning to progressively learn the object detectors in an easy-to-hard manner. Qian et al. \cite{qian2022incorporating} incorporated a difficulty evaluation score into training loss to alleviate the imbalance between easy and difficult samples.}

%feng2022weakly
%wang2023mol
%qian2023semantic

{The aforementioned studies improve the detection performance of image-level supervised RSOD methods. However, since image-level annotations cannot provide enough location and quantity information, these methods cannot achieve reasonable performance when applying to RSOD task (see Sec.~\ref{sec:experiments}). In this paper, we sacrifice little labor cost and focus on single pointly supervised RSOD. }

% For example, the state-of-the-art method MOL can achieve xxx on DOTA dataset, which is about xxx\% of the box-level supervised method Faster-RCNN 
 
%Due to the complex scenes in RS images and densely packed objects, existing proposal generators can not produce sufficient high-quality proposals for all objects. 

%cannot achieve reasonable performance when applying to RSOD task. 
% cannot meet the performance requirements in practice
%As discussed above, since image-level annotations cannot provide enough location and quantity information, image-level supervised detection methods cannot achieve reasonable performance when applying to RSOD task.

\subsection{Point Supervision in Vision Tasks}
Recently, point-level labels gradually attract research attention due to its similar labeling time and richer labeling information. Point-level supervision have been extensively investigated in many vision tasks, including object detection \cite{papadopoulos2017training, ren2020ufo, chen2022p2b}, semantic segmentation \cite{bearman2016s, qian2019weakly, wu2022deep}, instance segmentation \cite{laradji2020proposal, cheng2022pointly, liao2023attentionshift}, panoptic segmentation \cite{fan2022pointly}, localization \cite{ribera2019locating, song2021rethinking, yu2022object}, infrared small target segmentation \cite{ying2023mapping, li2023monte}, and so on. 

Wu et al. \cite{wu2022deep} proposed a deep bilateral filtering network (DBFNet) for single pointly supervised semantic segmentation, in which bilateral filter was introduced to enhance the consistency of features in smooth regions and enlarge the distance of features on different sides of edges. 
Cheng et al. \cite{cheng2022pointly} proposed a multi-pointly supervised instance segmentation method, named Implicit PointRend, that can generate parameters of the mask prediction function for each object. 
Fan et al. \cite{fan2022pointly} considered panoptic pseudo-mask generation as a shortest path searching puzzle, and used semantic similarity, low-level texture cues, and high-level manifold knowledge as traversing costs between adjacent pixels. 
Yu et al. \cite{yu2022object} proposed a coarse point refine (CPR) method for single pointly supervised object localization, and the CPR method can select semantic-correlated points around point labels and find semantic center points through MIL learning. 

{In object detection field, Papadopoulos et al. \cite{papadopoulos2017training} firstly introduced center-click annotation, in which the error distribution between two clicks is utilized to estimate object scales. Hence, two repetitive and independent center annotations are needed in their method. Different from that, our method try to generate pseudo boxes from single arbitrary point on the object mask. Ren et al. \cite{ren2020ufo} proposed a unified object detection framework (i.e., UFO$^2$) that can handle different forms of supervision (e.g., tags, points, scribbles and boxes) simultaneously. Different from handling different forms of supervision, the emphasis of our method is better generating pseudo boxes from single points based on the characteristics of RS objects.} Chen et al. \cite{chen2022p2b} proposed an MIL based single pointly supervised detection framework that can adaptively generate and refine proposals via multi-stage cascaded networks. In their method, proposal bags are generated through some fixed parameters that control the proposal scales, aspect ratios, shaking degrees and quantities. However, due to the challenges in RS field (as mentioned in Introduction), their method suffers a performance degradation when applying to RS images. In this paper, we focus on the special challenges of RSOD and explore single pointly supervised detection methods tailored for RS images.

\begin{figure*}[t]
	\centering
	\includegraphics[width=17cm]{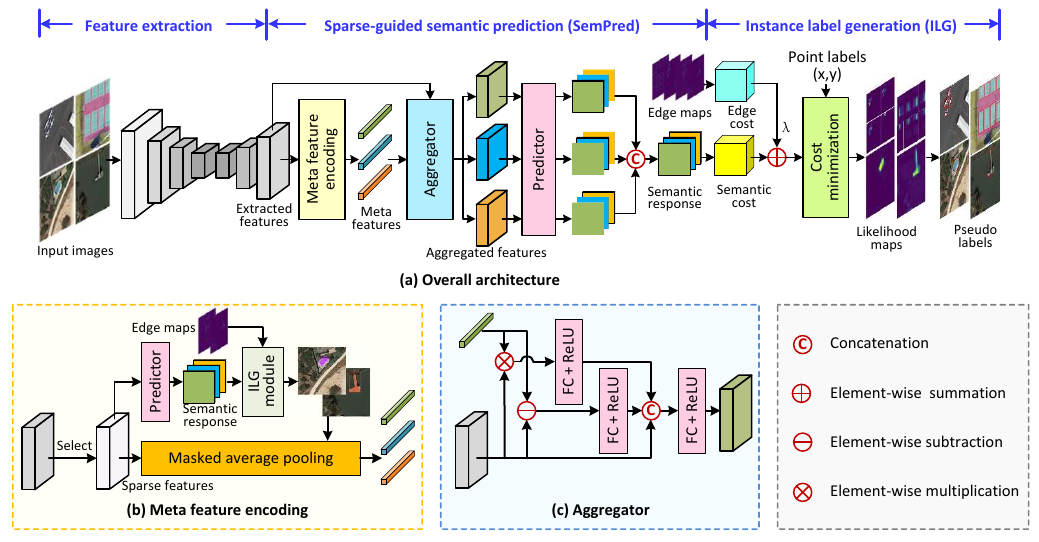}
%  \vspace{-0.2cm}
 	\caption{An overview of the proposed point label upgrader (PLUG), which is designed to transform point labels into pseudo boxes. Specifically, the feature extraction module extracts discriminative features from input images. Then, the sparse feature guided semantic prediction (SemPred) module takes the extracted features as its input and is responsible for the semantic response prediction. Finally, the instance label generation (ILG) module takes both the input images and the predicted response as its input to generate pseudo boxes.}
%   \vspace{-0.2cm}
	\label{fig_networks}
	%	\vspace{-0.2cm}
\end{figure*}

\section{Method}
\label{sec:methods}
In this section, we introduce the details of our method. We first introduce the architecture of the proposed point label upgrader (PLUG), which consists of the feature extraction module, the sparse feature guided semantic prediction (SemPred) module and the instance label generation (ILG) module (see Fig.~\ref{fig_networks}). Afterwards, we introduce the training losses of our PLUG.

\subsection{Feature Extraction}
In our method, ResNet \cite{ResNet} with FPN \cite{FPN} is used as the feature extraction module. The ResNet backbone extracts features of images of different scales, and FPN fuses the multi-scale features to balance the contents of semantic and structure information. Following \cite{yu2022object}, the P2 layer (with 8$\times$ down-sampling ratio) of FPN is used for subsequent processing.

\subsection{Sparse Feature Guided Semantic Prediction}
Taking the extracted features as input, the sparse feature guided semantic prediction (SemPred) module is responsible for obtaining the semantic response of objects, in which object regions are activated in the specific category layers. Besides, the SemPred module can reduce the difficulty of discriminative feature extraction on dense objects. Specifically, we observe that the pseudo boxes generated on sparse objects are of higher quality than those generated on dense objects (see Sec.~\ref{motivation} for details). Consequently, in our SemPred module, the general representation of sparse objects is used to enhance the extracted features, and thus improves the discriminative feature representation capability of our PLUG. The detailed architecture of the SemPred module is shown in Fig.~\ref{fig_networks}(a), which consists of three stages: meta feature encoding, feature aggregation and semantic prediction.

\textit{\textbf{1) Meta feature encoding.}} In this stage, the general representation (i.e., meta feature) of sparse objects is encoded from the extracted features. As shown in Fig.~\ref{fig_networks}(b), meta feature encoding takes the extracted features as input, and obtains sparse features by selecting the features of images with single object. Then, the sparse features are fed to a predictor and the ILG module to generate the pseudo labels of sparse objects. With the sparse features and the pseudo labels, masked average pooling is performed to obtain the feature representation of each sparse object. In order to obtain more representative and stable meta features, all the sparse representations in the dataset are averaged according to their categories. Finally, $C$ (the number of categories) meta features are obtained, each of which can represent the general information of objects in a specific category. 

\textit{\textbf{2) Feature aggregation.}} After obtaining $C$ meta features, $C$ aggregated features are generated in this stage by using meta features to enhance the extracted features. The architecture of our aggregator is shown in Fig.~\ref{fig_networks}(c). Specifically, for each meta feature, element-wise subtraction and multiplication are first performed. Then, the processed features are concatenated with the original feature to obtain the aggregated features. Note that, a fully-connected layer and a ReLU layer are used after each operation (i.e., subtraction, multiplication and concatenation).

%we follow \cite{zhang2021meta} to perform feature aggregation, which is illustrated in Fig.~\ref{fig_networks} (c). For each meta feature, element-wise subtraction and multiplication are performed to extract different information. Then, these two features and the original features are concatenated to obtain the aggregated features. Note that, a fully-connected layer followed by a ReLU layer are used after each operation (i.e., subtraction, multiplication and concatenation). Finally, $C$ aggregated features can be obtained, where $C$ is the number of categories. 

\textit{\textbf{3) Semantic prediction.}} For each aggregated feature, a predictor (composed of a Linear layer and a Sigmoid function) is used for semantic response prediction. Since the representations in meta features are category-aware, different aggregated features are expert in predicting objects in corresponding categories. Hence, the specific layer of the semantic response from different aggregated features are selected and concatenated to generate the final semantic response. It is worth noting that the predictor in different branches and in the meta feature encoding module share the same architecture and parameters.

Note that, in the SemPred module, meta feature encoding is performed in the training phase only. During inference, the meta features have been optimized and stored in advance, and thus the extracted features can be directly aggregated. 
In fact, the guidance of sparse objects can be considered as a self-distillation process \cite{gou2021kdsurvey}, where the sparse features are the \emph{teacher} and can transfer  knowledge (high-quality features) to the \emph{student}. With the guidance of sparse objects, the semantic response can be enhanced, and benefits the pseudo box generation in the following ILG module.

\subsection{{Instance Label Generation}}
After obtaining the semantic response, the ILG module is designed to generate pseudo box annotations. The core of this module is to assign each pixel to its most likely object or background. Based on the assignment results, we can obtain the bounding box of each object by finding the circumscribed rectangle of the corresponding pixels.

Specifically, let $\mathcal{L} = \{l_{0}, l_{1}, l_{2}, ..., l_{L}\}$ denote the set of instances, where $l_{0}$ denotes background and $\{l_{1}, l_{2}, ..., l_{L}\}$ denote $L$ objects.
Each pixel $p$ on the image will be assigned to an instance according to
\begin{equation}\label{eq_assign}	
	\mathcal{I}ns(p) = \mathop{\arg\min}_{l\in\mathcal{L}} \left\{\textit{Cost}(p, p_{l})\right\},
\end{equation}
where $p_{l}$ represents the point label of instance $l$ that contains both location and instance information. $\textit{Cost}(p, p_{l})$ denotes the cost between pixel $p$ and point label $p_{l}$. The core of the label assignment process in Eq.~\ref{eq_assign} is to find an instance with minimum cost for each pixel. 

The cost calculation between pixel $p$ and point label $p_{l}$ is formulated as a shortest path problem. {Specifically, we formulate the cost between $p$ and $p_{l}$ as the second curvilinear curve integral along a given path $\Gamma\in \{\Gamma_1,..., \Gamma_n\}$. That is,
\begin{equation}\label{eq_assign2}	
		\textit{Cost}(p, p_{l}) = 		
			\min\limits_{\Gamma\in \{\Gamma_1,..., \Gamma_n\}} \int_{\Gamma} \left(C^{sem}(\vec{z}) + \lambda C^{edge}(\vec{z})\right)d\vec{z},		
\end{equation}
where $C^{sem}(\cdot)$ and $C^{edge}(\cdot)$ represent the semantic-aware neighbor cost and edge-aware neighbor cost, respectively, and $\lambda$ is a hyper-parameter to balance these two terms \cite{fan2022pointly}. Specifically, $C^{sem}(\cdot)$ is the $L_2$ distance of the semantic response between two adjacent pixels. $C^{edge}(\cdot)$ is the $L_1$ distance of the edge map (generated by Sobel operator \cite{duda1973pattern}) between two adjacent pixels, which can help better distinguish the densely packed objects (see Sec.~\ref{effect_ilg}). 
Note that, $\textit{Cost}(p, p_{0})$ is manually set to a fixed threshold $\tau$ ($\tau = 0.5$ in our method) to assign pixels that are ``far from'' all the instances to the background. Besides, since there is no analytical solution to the integral in Eq.~\ref{eq_assign2}, we use the Dijkstra's algorithm to obtain its numerical solution.}
% we use the Dijkstra's method \cite{dijkstra2022note} to solve this shortest path problem.

\subsection{Losses}
In the proposed PLUG, the ILG module is parameter-free, and the training process is only performed on the SemPred module. The losses to train the SemPred module have three parts including positive loss, negative loss and color prior loss. 

\textit{\textbf{1) Positive loss.}}
Since point labels can provide accurate supervision on the annotated locations, we set these labeled pixels as positive samples, and design a positive loss to optimize the SemPred module to generate correct predictions on these positions. The positive loss is designed based on the standard focal loss \cite{RetinaNet}:
\begin{equation}\label{eq_posloss}
	\begin{aligned}
	 \mathcal{L}_{pos} = &- \frac{1}{N_{pos}}\sum_{j=1}^{N_{pos}} \sum_{i=1}^{C}~[ y_{ji}(1-y^{'}_{ji})^{\gamma}log(y^{'}_{ji})\\
	 &+ (1-y_{ji})y_{ji}^{'\gamma}log(1-y^{'}_{ji}) ],
\end{aligned}
\end{equation}
where $N_{pos}$ and $C$ denote the total number of positive samples and categories, respectively. $y$ and $y^{'}$ represent the groundtruth category label and the prediction scores, respectively. We follow the \textcolor{black}{general} settings in \cite{RetinaNet} to set $\gamma$ to 2.

\textit{\textbf{2) Negative loss.}} In PSOD, only objects are labeled by single points, while the background regions are not annotated. Consequently, single point annotations cannot provide sufficient supervision on background. In our method, we follow this basic setting in PSOD and propose an approach to provide supervision on the background regions. Specifically, we suppose that background pixels are dominant in amount in the unlabeled region, and then coarsely set all the unlabeled pixels as negative samples. Based on the coarse negative samples, we design a negative loss to enforce our model to better distinguish objects and background, i.e.,
\begin{equation}\label{eq_negloss}
	\begin{aligned}
		\mathcal{L}_{neg} = - \frac{1}{N_{neg}}\sum_{j=1}^{N_{neg}} \sum_{i=1}^{C} (1-y_{ji})y_{ji}^{'\gamma}log(1-y^{'}_{ji}),
	\end{aligned}
\end{equation}
where $N_{neg}$ is the number of negative samples. 

\textit{\textbf{3) Color prior loss.}}
We follow \cite{fan2022pointly} to introduce a color prior loss, which can encourage adjacent pixels with similar colors be classified to the same category, and enhance the prediction stability of our SemPred module. The color prior loss is formulated as
\begin{equation}\label{eq_colorloss}
	\begin{aligned}
		\mathcal{L}_{\text {col}}=-\frac{1}{Z} \sum_{i=1}^{H W} \sum_{j \in \mathcal{N}(i)} A_{i, j} \log y_{i}^{'T} y_{j}^{'}.
	\end{aligned}
\end{equation}
where $y_{i}^{'}$, $y_{j}^{'}$ denote the category prediction scores of the $i^{th}$ and $j^{th}$ pixels, respectively. $A_{i,j}$ is the color prior affinity, and is obtained by thresholding the pixel similarity computed in the LAB color space (with a threshold of 0.3). $\mathcal{N}(i)$ is the set of neighbor pixel indices of $i$. $Z =\sum_{i=1}^{H W} \sum_{j \in \mathcal{N}(i)} A_{i, j}$ is the normalization factor.
% With the above loss functions, our PLUG can be well optimized and generate pseudo bounding boxes in an effective manner.

\textcolor{black}{In summary, the overall loss is the weighted summation of the above three losses, i.e.,}
\begin{equation}\label{eq_overallloss}
	\begin{aligned}
		\mathcal{L}_{\text {all}}=\mathcal{L}_{\text {pos}} + \alpha_{1} \mathcal{L}_{\text {neg}} + \alpha_{2} \mathcal{L}_{\text {col}},  
	\end{aligned}
\end{equation}
\textcolor{black}{where $\alpha_{1}$, $\alpha_{2}$ are two hyperparameters to balance different terms. In this paper, $\alpha_{1}$ and $\alpha_{2}$ are set to ${N_{neg}}/{N_{pos}}$ and 1, respectively. With the well designed loss function, our PLUG can be well optimized and generate pseudo bounding boxes in an effective manner.}
\definecolor{Ocean}{RGB}{129,194,234}
\begin{table*}
	
	\begin{center}
		{
		\caption{Average precision scores achieved by different detection methods on the DOTA dataset. Here, def-DETR represents Deformable DETR.}
%   \vspace{-0.2cm}
		\scriptsize
		\label{tbl_comparisontest}
		\renewcommand\arraystretch{1.3}
	\setlength{\tabcolsep}{1.2mm}{
			\begin{tabular}{rcc ccccc ccccc ccccc cc}		
				\hline
				\multirow{2}{*}{Method}&\multirow{2}{*}{Supervision}&\multirow{2}{*}{Backbone}&\multicolumn{15}{c}{Categories} &\multirow{2}{*}{\textit{mAP}$_\textit{50}$} \\
				\cline{4-18}
				 & & & PL &BD& BR & GTF & SV& LV & SH & TC &BC & ST & SBF & RA & HB & SP & HC& \\
				\hline
				\rowcolor{gray!5} 
    WSDDN \cite{bilen2016weakly}&\textit{Image}&VGG16 &0.003&0.009&0.000&0.005&0.000&0.001&0.000&0.003&0.000&0.000&0.013&0.000&0.010&0.010&0.000&0.004\\
				\rowcolor{gray!5} 
    WSDDN \cite{bilen2016weakly} &\textit{Image}&ResNet50  &0.014&0.064&0.001&0.013&0.021&0.030&0.016&0.034&0.004&0.025&0.019&0.053&0.011&0.044&0.004&0.023\\
			    \rowcolor{gray!5} 
       OICR \cite{tang2017multiple}&\textit{Image}&VGG16  &0.007&0.100&0.000&0.116&0.037&0.101&0.023&0.089&0.000&0.056&0.145&0.000&0.042&0.036&0.000&0.050\\
				\rowcolor{gray!5} 
    OICR \cite{tang2017multiple} &\textit{Image}&ResNet50  &0.047&0.104&0.007&0.042&0.022&0.061&0.022&0.068&0.031&0.044&0.096&0.102&0.061&0.047&0.016&0.051\\
				\rowcolor{gray!5} 
    OICR-FR \cite{tang2017multiple} &\textit{Image}&ResNet50  &0.042&0.038&0.000&0.002&0.075&0.301&0.037&0.077&0.011&0.132&0.033&0.159&0.050&0.120&0.001&0.072\\
				\hline
			\rowcolor{gray!10} 
		FCOS \cite{FCOS}&\textit{Box}&ResNet50  &0.800&0.504&0.296&0.212&0.603&0.796&0.821&0.914&0.452&0.612&0.407&0.460&0.751&0.213&0.313&0.544\\
        \rowcolor{gray!10} 
\text{def-DETR} \cite{zhu2020deformable} &\textit{Box}& ResNet50&0.799&0.576&0.377&0.491&0.600&0.772&0.843&0.924&0.414&0.624&0.457&0.396&0.721&0.455&0.324&0.577 \\
    				\rowcolor{gray!10} 
        Faster-RCNN \cite{FasterRCNN}&\textit{Box}&ResNet50  &0.850&0.665&0.435&0.587&0.588&0.831&0.833&0.933&0.493&0.634&0.590&0.589&0.791&0.534&0.373&0.648\\
				\hline
				\rowcolor{gray!10}
    P2BNet-FR \cite{chen2022p2b}& \textit{Point}&ResNet50 & 0.061 & 0.063 & 0.111 & 0.260 & 0.266 & 0.066 & 0.368 & 0.016 & 0.051 & 0.270 & 0.049 & 0.272 & 0.105 & 0.386 & 0.001 & 0.156\\
               \rowcolor{gray!10} 
               P2BNet-FR* \cite{chen2022p2b}& \textit{Point}&ResNet50 & 0.016 & 0.002 & 0.118 & 0.168 & \textbf{0.397} & 0.073 & 0.246 & 0.017 & 0.190 & \textbf{0.465} & 0.009 & 0.518 & 0.060 & 0.358 & \textbf{0.140} & 0.185\\
    \rowcolor{green!15} 
    PLUG-FCOS(ours) &\textit{Point}&ResNet50 &0.353&0.340&0.226&0.111&0.296&\textbf{0.685}&\textbf{0.603}&\textbf{0.874}&\textbf{0.246}&0.455&0.192&0.468&0.349&0.171&0.039&0.360 \\
    \rowcolor{green!15}
    \text{PLUG-def-DETR} (ours) &\textit{Point}&ResNet50&0.250&0.398&0.241&0.166&0.288&0.614&0.547&0.795&0.090&0.383&0.160&0.345&0.227&0.353&0.047&0.322 \\
    				\rowcolor{green!15} 
    PLUG-FR (ours)& \textit{Point}&ResNet50  &\textbf{0.509}&\textbf{0.543}&\textbf{0.291}&\textbf{0.284}&0.248&0.672&0.436&\textbf{0.874}&0.214&0.462&\textbf{0.360}&\textbf{0.543}&\textbf{0.438}&\textbf{0.446}&0.086&\textbf{0.427} \\
				\hline
		\end{tabular}}
		\begin{tablenotes} %添加此处
			\item * means that P2BNet is optimized in a two-stage cascaded manner.
		\end{tablenotes} }%添加此处
%   \vspace{-0.3cm}
	\end{center}
\end{table*}

\begin{figure*}[t]
	\centering
	\includegraphics[width=17cm]{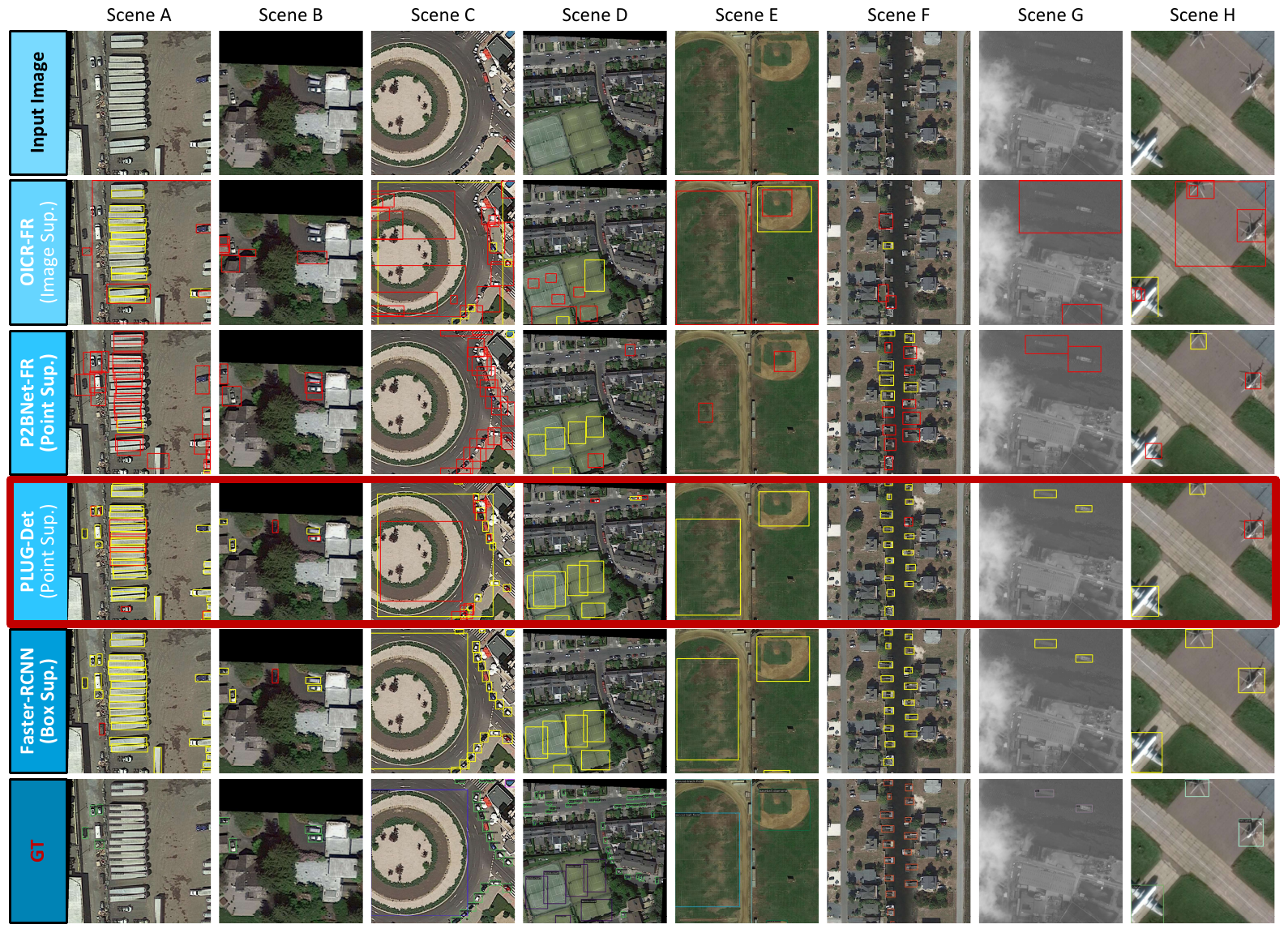}
%  \vspace{-0.2cm}
	\caption{Qualitative results obtained by different object detection methods on the DOTA validation set. The correctly detected results are marked by yellow boxes, and the falsely detected results are marked by red boxes. Gradually darker colors represent stronger supervision.}
%  \vspace{-0.2cm}
	\label{fig_detvisual2}
	%	\vspace{-0.2cm}
\end{figure*}
\section{Experiments}
\label{sec:experiments}
In this section, we firstly introduce the datasets and implementation details, and then combine the proposed PLUG with Faster-RCNN \cite{FasterRCNN} to develop a PSOD method (i.e., PLUG-Det). Afterwards, we compare  PLUG-Det with image-level, point-level and box-level supervised object detection methods. Moreover, we conduct ablation studies and make deep analyses to validate the effectiveness of our method. Finally, we develop a PLUG-Seg network by combing PLUG with Mask-RCNN \cite{MaskRCNN}, and conduct experiments to show the potential of our method in single pointly supervised instance segmentation (PSIS).
\subsection{Datasets and Implementation Details}
\subsubsection{\textbf{Datasets}} 
To verify the effectiveness of our method, we conduct extensive experiments on the DOTA-v1.0 dataset \cite{DOTA}, which contains 2806 large-scale RS images with 15 object categories, including plane (PL), baseball diamond (BD), bridge (BR), ground track field (GTF), small vehicle (SV), large vehicle (LV), ship (SH), tennis court (TC), basketball court (BC), storage tank (ST), soccer ball field (SBF), roundabout (RA), harbor (HB), swimming pool (SP) and helicopter (HC). Objects in the DOTA dataset are labeled with box annotations. Since the iSAID dataset \cite{waqas2019isaid} contains the corresponding mask labels of objects in the DOTA dataset, we randomly selected a point on the mask of each object as the groundtruth point label. We used the training set and validation set for model development and performance evaluation, respectively. Due to hardware memory limitation, we cropped the original images into 512$\times$512 patches with 128 overlapped pixels, and used the cropped patches for training and inference. In the training phase, random flip was used for data augmentation.

\subsubsection{\textbf{Implementation Details}}
	We implemented our method based on the MMDetection \cite{mmdetection} toolbox with an NVIDIA RTX 3090Ti GPU. {The training of our PLUG-Det method consists of three stages: the training of PLUG, the inference of PLUG and the training of existing detector (e.g., Faster-RCNN). In the first stage, the learning rate was initially set to 0.001 and decreased by a factor of 0.1 at the 8$^{\text{th}}$ and 11$^{\text{th}}$ epoch, respectively. We trained our PLUG for totally 12 epochs with a batch size of 8. Besides, we used the stochastic gradient descent (SGD) algorithm \cite{robbins1951stochastic} for optimization. In the second stage, pseudo boxes of the training set were obtained by performing inference using the trained PLUG. In this stage, the batch size was set to 1. 
	In the third stage, we adopted existing detector by default without modifying its hyper-parameters. Taking Faster-RCNN with ResNet50 as an example, the learning rate was initially set to 0.005, and the optimizer was SGD with 1$\times$ training schedule. Other training settings were kept as the default values in MMDetection \cite{mmdetection}. The training time of the three stages are 4.8, 6 and 3.1 hours, respectively. The total training time is the summation of the time spent in each stage, and is about 14 hours.}

\subsubsection{\textbf{Evaluation Metrics}}
We used $\textit{mIoU}$ between generated pseudo boxes and groundtruth boxes to evaluate the performance of PLUG. Besides, $\textit{mIoU}_\textit{s}$, $\textit{mIoU}_\textit{m}$ and $\textit{mIoU}_\textit{l}$ were used as the indicators to evaluate the quality of pseudo boxes on small, medium and large objects, respectively. 
Moreover, we evaluated the performance of PLUG-Det and its variants by reporting the \textit{mAP}$_\textit{50}$ (averaged over $\textit{IoU}$ values with the threshold being set to 0.5) for all categories and the $\textit{AP}_\textit{50}$ for each category. Similarly, \textit{mAP}$_\textit{s}$, \textit{mAP}$_\textit{m}$ and \textit{mAP}$_\textit{l}$ were used to evaluate the detection performance on small, medium and large objects, respectively.

\subsection{Comparison to the State-of-the-art Methods} 
In this subsection, we use the pseudo boxes generated by different methods to train a Faster-RCNN detector, and compare the detection performance of our PLUG-Det with existing image-level supervised and single pointly supervised detection methods. Moreover, Faster-RCNN with groundtruth box-level supervision is also included to provide upper bound results for reference.

Table~\ref{tbl_comparisontest} shows the \textit{AP}$_\textit{50}$ values achieved by different detection methods. It can be observed that image-level supervised detectors (i.e., WSDDN \cite{bilen2016weakly}, OICR \cite{tang2017multiple}, OICR-FR \cite{tang2017multiple}) achieve very low detection accuracy. Compared to those detectors, PSOD methods achieve better detection performance due to the extra coarse position and quantity information introduced by point annotations. Specifically, P2BNet-FR achieves an \textit{mAP}$_\textit{50}$ score of 0.156, and can further achieve a 0.029 improvement with a two-stage cascaded optimization pipeline. {In contrast, our PLUG-FR achieves an \textit{mAP}$_\textit{50}$ score of 0.423, which significantly outperforms P2BNet-FR. The experimental results demonstrate the superiority of our method as compared to the MIL-based methods. It is worth noting that Faster-RCNN developed under groundtruth box supervision can achieve an \textit{mAP}$_\textit{50}$ score of 0.648. That is, our PLUG-FR can achieve 65.3\% of the performance of box-level supervised Faster-RCNN \cite{FasterRCNN}, but with a 18$\times$ reduction in annotation cost. }
	
{Besides, our method can generalize to different downstream detectors. We additionally use the one stage detector FCOS \cite{FCOS} and the Transformer based detector Deformable DETR \cite{zhu2020deformable} to validate the generalization capability of our method. As shown in Table~\ref{tbl_comparisontest}, PLUG-FCOS and PLUG-Deformable DETR can achieve 0.360 and 0.322 in terms of $\textit{mAP}_\textit{50}$, and are 66.2\% and 55.8\%  of the performance of each fully supervised detectors, respectively. The consistent performance ratios compared to respective fully supervised detectors demonstrate the generality of our method.}

Figure~\ref{fig_detvisual2} shows the qualitative results on eight typical scenes achieved by different detection methods. {It can be observed that our PLUG-Det can achieve better detection performance than other state-of-the-art image-level supervised and single pointly supervised detectors, especially on challenging scenes. Specifically, image-level supervised detectors (e.g., OICR-FR) may bring false alarms (e.g., Scene C) and miss detection (e.g., Scene F) due to its insufficient supervision. Besides, single pointly supervised detector P2BNet-FR has worse scale and aspect ratio adaptability compared with our method. For example, the vehicles in Scene A with large aspect ratios cannot be correctly detected by P2BNet-FR, but can be better detected by our method.}
	   
%	 can be accurately detected by our method.}
%but our method can accurately detect most of them
%	For example, objects in Scene E are in different scales, and the bounding boxes predicted by the P2BNet-FR are smaller than the ground-truth boxes, which results in miss detection. In contrast, our method can generate more accurate bounding boxes on objects with varied scales. In scene A, the vehicles with large aspect ratios cannot be correctly detected by OICR-FR and P2BNet-FR. In contrast, our method can detect most of the vehicles. 

%It can be observed that the scale adaptability of P2BNet-FR is worse than our method. For example, in Scene E, the bounding boxes predicted by the P2BNet-FR are smaller than the ground-truth boxes, which results in miss detection. In contrast, our method can generate more accurate bounding boxes on objects with varied scales. Besides, the aspect ratio adaptability of P2BNet-FR is also worse than our method. For example, in Scene A, the vehicles with large aspect ratios cannot be correctly detected by P2BNet-FR. In contrast, our method can detect most of the vehicles. 

\begin{table*}
	\begin{center}
		\caption{Comparison of the pseudo box quality and detection performance achieved by different backbones. Here, \#Param represents the number of parameters, and FLOPs is calculated with a 512$\times$512 input image.}
%   \vspace{-0.2cm}
    \label{tbl_feaextract}
		\renewcommand\arraystretch{1.3}
		\setlength{\tabcolsep}{3mm}{
		\begin{tabular}{ccc cccc cccc}
			\hline
			\multirow{2}{*}{backbone}&\multirow{2}{*}{FLOPs}&\multirow{2}{*}{\#Param}&\multicolumn{4}{c}{Pseudo box quality}&\multicolumn{4}{c}{Detection performance}\\
   \cline{4-11}
   &&&$\textit{mIoU}$ &	$\textit{mIoU}_\textit{s}$ &$\textit{mIoU}_\textit{m}$&$\textit{mIoU}_\textit{l}$&\textit{mAP}$_\textit{50}$ &\textit{mAP}$_\textit{s}$ &\textit{mAP}$_\textit{m}$&\textit{mAP}$_\textit{l}$\\
			\hline
			ResNet18&12.17 G&13.37 M&0.531&0.524&0.551&0.508&0.412&0.330&0.457&0.262\\
			ResNet50&24.88 G&26.32 M&0.549&0.539&0.576&0.533&0.427&0.329&0.474&0.338\\
			ResNet101&44.36 G&45.31 M&\textbf{0.558}&\textbf{0.548}&\textbf{0.584}&\textbf{0.564}&\textbf{0.436}&\textbf{0.338}&\textbf{0.490}&\textbf{0.384}\\
			\hline
		\end{tabular}
	}
	\end{center}
%  \vspace{-0.2cm}
\end{table*}

\subsection{Ablation Study}\label{sec_abl}
In this section, we conduct ablation studies to validate the effectiveness of our method.

\subsubsection{\textbf{Investigation of the Feature Extraction Module}}

\begin{figure}[t]
	\centering
	\includegraphics[width=8cm]{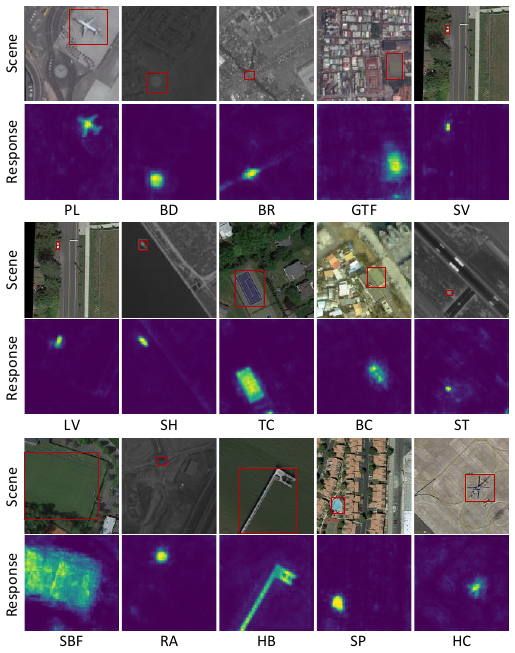}
%  \vspace{-0.2cm}
	\caption{The semantic response of images predicted by the SemPred module. Here, the layer of the corresponding category is visualized.}
%  \vspace{-0.2cm}
	\label{fig_dsp}
\end{figure}

We use ResNet \cite{ResNet} with FPN \cite{FPN} as the feature extraction module of our PLUG. Here, we compare the performance of our feature extraction module with different backbones (i.e., ResNet18, ResNet50 and ResNet101). We first evaluate the quality of generated pseudo boxes on the training set. As shown in Table~\ref{tbl_feaextract}, our PLUG can achieve $\textit{mIoU}$ scores of 0.531, 0.549 and 0.558 with ResNet18, ResNet50 and ResNet101 backbones, respectively. We also evaluate the downstream detection performance on the validation set. As shown in Table~\ref{tbl_feaextract}, our PLUG-Det achieves an \textit{mAP}$_\textit{50}$ score of 0.436 with ResNet101, which is higher than the \textit{mAP}$_\textit{50}$ scores with ResNet18 and ResNet50. That is because, the ResNet101 backbone is deeper and can extract more discriminative features. {However, compared to ResNet18 and ResNet50, using ResNet101 as backbone introduces larger model size (1.82$\times$ of ResNet50) and higher FLOPs (1.78$\times$ of ResNet50). Consequently, we use ResNet50 as the backbone to achieve a good balance between accuracy and efficiency.}

\begin{figure}[t]
	\centering
	\includegraphics[width= 8.5 cm]{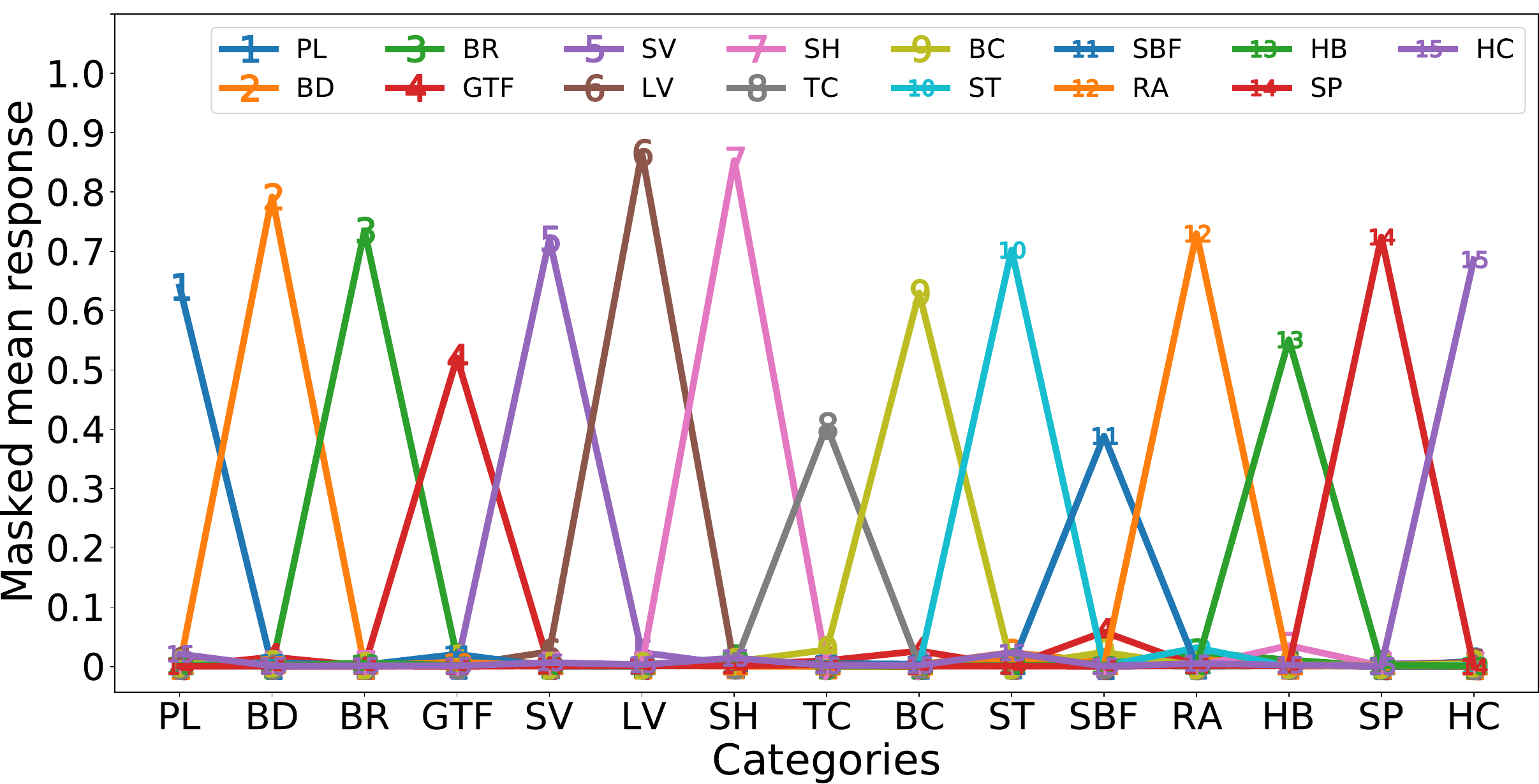}
%  \vspace{-0.2cm}
	\caption{The distribution of masked mean response in different categories.}
%  \vspace{-0.2cm}
	\label{fig_dsp2}
\end{figure}

\subsubsection{\textbf{Effectiveness of the SemPred Module}}

The SemPred module utilize meta features of sparse objects to aggregate the extracted features, and use the aggregated features for semantic response prediction. We conduct experiments to validate the effectiveness of the SemPred module and its key components.

Semantic response visualization. The semantic response prediction contains two potential tasks, including the recognition of objects from background and the discrimination among categories. In this part, we validate the effectiveness of the SemPred module on these two tasks, respectively. First, we validate the object recognition capability by visualizing the predicted semantic response maps. As shown in Fig.~\ref{fig_dsp}, objects of different categories can be well distinguished from background, and the response regions basically fit object shapes. Second, we validate the category discrimination capability of the SemPred module by visualizing the variation of masked mean response\footnote{Masked mean response denotes the average value of response map on the groundtruth mask of each object.} on different category layers.  As shown in Fig.~\ref{fig_dsp2}, each object is only strongly activated on a single category layer. These results clearly demonstrate the effectiveness of the SemPred module on recognizing and classifying objects from backgrounds. %It can be further observed in Fig.~\ref{fig_dsp2} that the response varies among classes. For example, the masked average response of large vehicle (LV) is about 0.9, while the response of tennis court (TC) is only about 0.4. %We intuitively think this is resulted by the varied scale distribution of objects.

% These visualization results clearly demonstrate the effectiveness of our SemPred module on semantic recognition. 

% of objects. It is worth noting that for each object, only the response layer of its category is visualized for better visual effects. As shown in Fig.~\ref{fig_dsp}, the foreground of different objects can be well distinguished, and the response regions basically fits object shapes. These visualization results directly demonstrate the effectiveness of our SemPred module on semantic recognition. Based on the visualization on Fig.~\ref{fig_dsp}, we additionally visualize the variation of masked mean response on different category layers of objects to validate the class discrimination capability of the DSP module. Note that, masked mean response refers to the average value of response on the groundtruth mask for each object. As illustrated in Fig.~\ref{fig_dsp2}, each object is only strongly activated on a single category layer. This validate the effectiveness of the SemPred module on classifying objects in different categories. 

%Besides, as shown in Fig.~\ref{fig_dsp} and Fig.~\ref{fig_dsp2}, the masked average response of objects varies considerably according to categories. For example, the masked average response of large vehicle (LV) is about 0.9, while for tennis court (TC) the according response is only about 0.4. We intuitively think this is resulted by the varied scale distribution of objects.
 
\begin{table*}[t]
	\begin{center}
		{\caption{Comparison of the pseudo box quality and detection performance achieved by our PLUG with different semantic prediction modules. Note that, the vanilla and SemPred module represent the method without and with performing sparse feature guidance, respectively.}
			%     \vspace{-0.2cm}
			\label{tbl_sga}
			%		\scriptsize
			\renewcommand\arraystretch{1.3}
			\begin{tabular}{ccccc cccc}
				\hline
				\multirow{2}{*}{semantic prediction module}&\multicolumn{4}{c}{Pseudo box quality}&\multicolumn{4}{c}{Detection performance}\\
				\cline{2-9}
				&$\textit{mIoU}$ & $\textit{mIoU}_\textit{s} $ &$\textit{mIoU}_\textit{m}$&$\textit{mIoU}_\textit{l}$ & \textit{mAP}$_\textit{50}$ & \textit{mAP}$_\textit{s}$ & \textit{mAP}$_\textit{m}$ & \textit{mAP}$_\textit{l}$ \\
				\hline
				vanilla&0.497&0.494&0.512&0.457&0.356&0.292&0.401&0.215\\
				SemPred&0.549&0.539&0.576&0.533&0.427 &0.329&0.474&0.338\\
				\hline
		\end{tabular}}%}
		%   \vspace{-0.2cm}
	\end{center}
\end{table*}

\begin{figure*}[t]
	\centering
	\includegraphics[width= 16 cm]{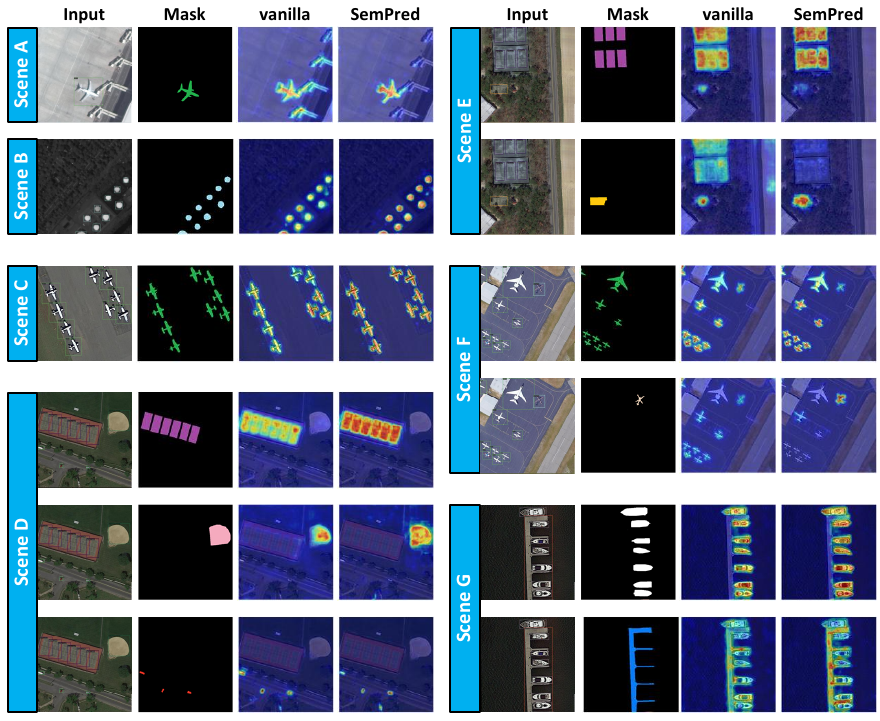}
	\vspace{-0.2cm}
	\caption{The heatmaps of specific response layers produced by our SemPred module with and without performing sparse feature guidance (SFG).}
	%  \vspace{-0.2cm}
	\label{fig_sga}
\end{figure*}

Sparse feature guidance. In the SemPred module, the general representations of sparse objects are used to aggregate the extracted features from backbones. To validate the sparse feature guidance scheme, we replaced the SemPred module with a vanilla predictor (a Linear layer followed by a Sigmoid function), and developed a variant (i.e., ``vanilla" in Table~\ref{tbl_sga} ) of PLUG without the guidance of sparse objects. As shown in Table~\ref{tbl_sga}, the $\textit{mIoU}$ score is improved from 0.497 to 0.549 when sparse feature guidance is performed, and the \textit{mAP}$_\textit{50}$ value of our PLUG-Det is also improved from 0.356 to 0.423 correspondingly. It demonstrates that the proposed sparse feature guidance scheme can improve the quality of generated pseudo boxes, and thus benefits to the downstream detection performance. {Moreover, we compare the semantic response maps produced by our PLUG and its variant (vanilla and SemPred). We can draw the following conclusions from Fig.~\ref{fig_sga}:}

\begin{itemize}
	\item The sparse feature guidance scheme can improve the recognition capability of our PLUG on confusing background. As shown in Scene A, the plane (PL) and the boarding bridges are similar in color space. With the guidance of sparse features, our PLUG can better distinguish objects from background.

	\item The sparse feature guidance scheme can improve the recognition capability of our PLUG on dense objects. {For densely packed objects of the same category (e.g., Scenes B and C), some objects are weakly activated when sparse feature guidance is not performed. In contrast, by performing sparse feature guidance, the features of each object can be enhanced, and the intra-class instance recognition performance is improved. Besides, sparse feature guidance can also improve the recognition capability of our PLUG on densely packed objects of different categories (e.g., the ships (SH) and harbor (HB) in Scene G).}
	
	\item The sparse feature guidance scheme can enhance the capability of our PLUG to distinguish objects in different categories but with similar appearance. As shown in Scene E, the tennis court (TC) and basketball court (BC) have similar appearance, and our PLUG without sparse feature guidance cannot distinguish them and produces falsely mixed response.  Since category-aware meta features are used to aggregate the extracted features, the enhanced features have stronger category characteristics. Consequently, our PLUG with sparse feature guidance can effectively handle this mixed response issue and can well distinguish similar objects. 
\end{itemize}

\begin{figure}
	\centering
	\includegraphics[width= 7 cm]{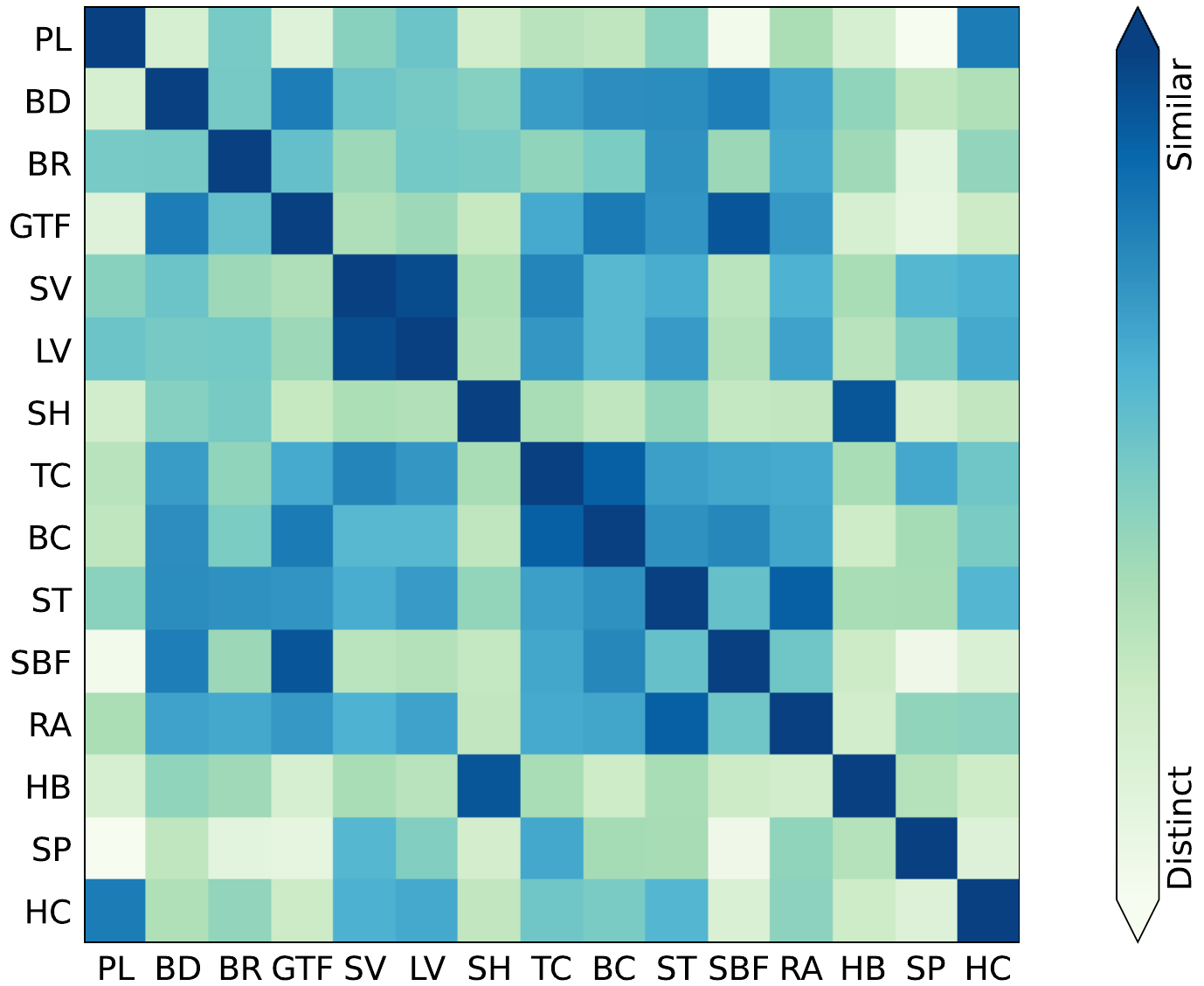}
	%  \vspace{-0.2cm}
	\caption{The cosine similarities between different pairs of representations in meta features. Here, darker colors indicate larger values (i.e., higher similarity).}
	%  \vspace{-0.2cm}
	\label{fig_prototype}
	%	\vspace{-0.2cm}
\end{figure}

Cross-category correlation of meta features. Meta features are the general representation of objects in different categories. Here, we visualize the cosine similarity map between each pair of meta features to investigate their correlation. As shown in Fig.~\ref{fig_prototype}, apart from the elements on the diagonal, there are still some pairs of meta features (e.g., large vehicle (LV) vs. small vehicle (SV), plane (PL) vs. helicopter (HC), basketball court (BC) vs. tennis court (TC)) highly correlated due to the similar appearance of the objects. This observation is consistent with the visualization results in Fig.~\ref{fig_sga}, and can demonstrate the effectiveness of the usage of meta features. 

\subsubsection{\textbf{Effectiveness of the Edge-aware Neighbor Cost}}
\label{effect_ilg}

\begin{figure*}[t]
	\centering
	\includegraphics[width=16cm]{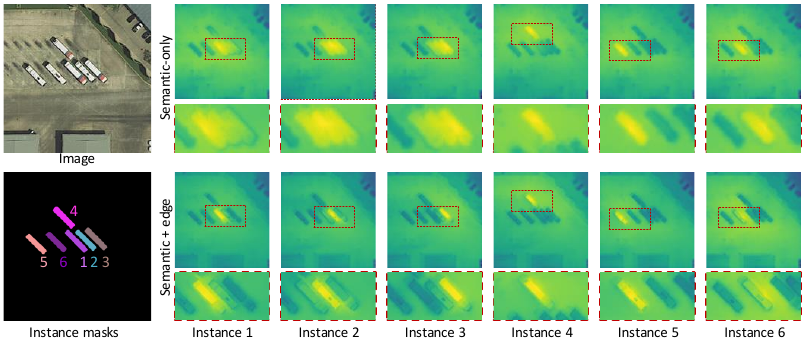}
	%  \vspace{-0.2cm}
	\caption{The likelihood maps generated by the ILG module with and without using the edge-aware neighbor cost. Note that, we visualize $P_{map}=1- C_{map}$ for better visual analyses, where $C_{map}$ is the cost map for each labeled point, and the values on the cost map can represent the costs from each pixel to the labeled point. Consequently, $C_{map} \in H \times W \times N$, where H and W are the height and width of images and $N$ is the number of objects in the image. Based on $C_{map}$, $P_{map}$ can represent the likelihood of each pixel belonging to a specific instance, and thus can more intuitively show the diffusion of labeled points. }
	%  \vspace{-0.2cm}
	\label{fig_ilg}
	%	\vspace{-0.2cm}
\end{figure*}

In this subsection, we validate the effectiveness of the edge-aware neighbor cost in the ILG module. Figure~\ref{fig_ilg} shows the likelihood maps $P_{map}=1- C_{map}$ with and without using edge-aware neighbor cost on an example scene, where the values represent the likelihood of a pixel belonging to a specific instance. It can be observed that densely packed adjacent instance can not be well distinguished without using edge-aware neighbor cost. That is because, the semantic-aware neighbor cost encourages the labeled points diffusing to the adjacent semantic-similar areas, and tends to consider the densely packed objects as a single instance. When the edge-aware neighbor cost is introduced, the diffusion of labeled points can stop at the boundaries, and these densely packed objects can be better distinguished. 

Note that, the value of $\lambda$ in Eq.~\ref{eq_assign2} should be properly set to ensure preferable growth from point labels. We compare the quality of pseudo boxes and the detection performance with respect to different $\lambda$ values. As shown in Table~\ref{tbl_ilg}, when $\lambda$ is set to 0.5, our PLUG can generate pseudo boxes of the highest quality, and our PLUG-Det can achieve the best detection performance. Consequently, we set $\lambda$ to 0.5 to balance the semantic-aware and edge-aware neighbor cost.

\begin{table}
	\begin{center}
		\caption{Comparison of the pseudo box quality and detection performance achieved by our PLUG with different $\lambda$ values. Best results are in bold faces.}
		%   \vspace{-0.2cm}
		\label{tbl_ilg}
		\renewcommand\arraystretch{1.3}
		\setlength{\tabcolsep}{1.55mm}{
			\begin{tabular}{cccc cccc c}
				\hline
				\multirow{2}{*}{$\lambda$}&\multicolumn{4}{c}{Pseudo box quality}&\multicolumn{4}{c}{Detection performance}\\
				\cline{2-9}
				&$\textit{mIoU}$&$\textit{mIoU}_\textit{s}$&$\textit{mIoU}_\textit{m}$&$\textit{mIoU}_\textit{l}$&\textit{mAP}$_\textit{50}$&\textit{mAP}$_\textit{s}$&\textit{mAP}$_\textit{m}$&\textit{mAP}$_\textit{l}$\\
				\hline
				0&0.497&0.473&0.552&0.548&0.405&0.305&0.461&0.340\\
				0.5&\textbf{0.549}&0.539&\textbf{0.576}&\textbf{0.533}&\textbf{0.427}&\textbf{0.329}&0.474&\textbf{0.338}\\
				1.0&0.547&\textbf{0.541}&0.567&0.528&0.425&0.327&0.467&0.319\\
				1.5&0.542&0.536&0.559&0.523&0.426&0.322&\textbf{0.475}&0.330\\
				2.0&0.517&0.389&0.547&0.552&0.422&0.328&0.473&0.335\\
				\hline
		\end{tabular}}
	\end{center}
	%  \vspace{-0.2cm}
\end{table}

\subsubsection{\textbf{Effectiveness of Losses}}
In this subsection, we conduct ablation studies to validate the effectiveness of the proposed losses. As shown in Table~\ref{tbl_loss}, our PLUG can only achieve an $\textit{mIoU}$ of 0.318 when the positive loss is used only. That is because, the background can not be considered in the training process, and thus degrades the recognition capability of our PLUG to distinguish objects and background. When the negative loss is introduced, both the quality of pseudo boxes and the detection performance are significantly improved. Moreover, applying the color prior loss can further introduce a 0.051 improvement of $\textit{mIoU}$ and a 0.006 improvement of $\textit{mAP}_\textit{50}$. The experimental results demonstrate the effectiveness of the proposed losses.

\begin{table}
	\begin{center}
		\caption{Comparison of the pseudo box quality and detection performance achieved by our PLUG with different losses.}
		%   \vspace{-0.2cm}
		\label{tbl_loss}
		\renewcommand{\arraystretch}{1.3}
		\setlength{\tabcolsep}{3.5mm}{
			\begin{tabular}{cccccc}
				\hline
				\multicolumn{3}{c}{Loss} & \multirow{2}{*}{$\textit{mIoU}$} & \multirow{2}{*}{\textit{mAP}$_\textit{50}$}\\
				\cline{1-3}
				positive&negative&color prior& &\\
				\hline
				$\checkmark$&&&0.318&0.175\\
				%			\hline
				$\checkmark$&$\checkmark$ &&0.498 &0.421\\
				$\checkmark$&$\checkmark$ &$\checkmark$ &\textbf{0.549}&\textbf{0.427}\\
				\hline
		\end{tabular}}
		%   \vspace{-0.2cm}
	\end{center}
\end{table}

\subsection{{Analyses of the Selecting Strategy of Point Labels}}

\begin{figure}[t]  
	\centering
	
	\includegraphics[width= 8.5 cm]{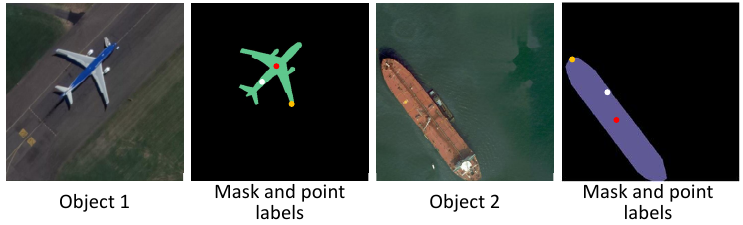}
	%  \vspace{-0.2cm}
	\caption{{Objects with point labels under different point selection strategies. The red points are the centers of masks, the yellow points are the corners of masks, and the white points are the randomly selected points on masks.}}
	%  \vspace{-0.3cm}
	\label{fig_pointselect}
	%	\vspace{-0.2cm}
\end{figure}

{In the preceding experiments, point labels were randomly selected from object masks. How will the locations of the selected points affect the performance? In this subsection, we implement three kinds of point labels, and conduct experiments to analyze their impacts on the quality of generated pseudo boxes and downstream detection performance. Specifically, we adopt three different labeling strategies, i.e., selecting the point in the center, selecting the point on the corner, and randomly selecting a point on the mask. Note that, since there is no clear definition about the corners of objects, we just selected the point (on the mask) that is farthest from the center point as its ``corner" label. Objects with different point labels are shown in Fig.~\ref{fig_pointselect}.}

%we implement the following three strategies: the centers, the corners and the randomly selected points on object masks. Objects with different point labels are shown in Fig.~\ref{fig_pointselect}.
%Note that, some centers may not belong to the objects. At this time, the point on the mask that is closest to the center is selected instead.
%%At this time, we select the points on the masks that are closest to the centers alternatively.
%Besides, since there is no clear definition about the corner of an object, we just select the point farthest from the center point as the corner label.}

 { Table~\ref{tbl_pointselect} shows the quality of pseudo boxes and the detection performance of our method with different point labels. It can be observed that our PLUG with center point labels can achieve the most superior results, which are 0.553 in terms of $\textit{mIoU}$ and 0.438 in terms of $\textit{mAP}_\textit{50}$. Besides, when the randomly selected points are used, the performance is slightly decreased (0.549 and 0.427 in terms of $\textit{mIoU}$ and $\textit{mAP}_\textit{50}$, respectively). Moreover, the corner labels result in a larger degree of performance degradation, in which $\textit{mIoU}$ and $\textit{mAP}_\textit{50}$ are decreased to 0.518 and 0.406, respectively.} %The marginally performance gaps validate the generality capability of our method to different point location.
  
%{Table~\ref{tbl_pointselect} shows the pseudo box quality and detection performance achieved by our method with different point labels. It can be observed that the quality of pseudo boxes generated by our PLUG can achieve 0.543, 0.553 and 0.549 $\textit{mIoU}$ with corners, centers and randomly selected points as labels, respectively. Besides, the PLUG-Det can achieve 0.434, 0.438 and 0.427 $\textit{mAP}_\textit{50}$ with different label selection strategies, separately. The comparative pseudo box qualities and detection performance demonstrate that our method can generalize to different point locations.}

{It is worth noting that the performance of our method with corner point labels is inferior than that with center and random point labels. That is because, the edge-aware neighbor cost used in the ILG module hinders the pixel diffusion of corner points. Specifically, the edge-aware neighbor cost is utilized to help stopping the diffusion of labeled points at boundaries, and thus prevent the labeled points from spreading towards the background areas (see Sec.~\ref{effect_ilg}). However, since the corner points are located on the boundaries of objects, the edge-aware cost may hinder the diffusion of the labeled points to the internal area of the object, as their paths pass through the edges. For example, as shown in the $P_{map}$ of the instance 6 in Fig.~\ref{fig_ilg}, the ILG module can recognize its correct regions with the semantic-aware cost only. However, when the edge-aware cost is introduced, the labeled points can only be diffused to background areas.}
%	Moreover, the performance achieved by our method with corner point labels is slightly lower than the center and random selected point labels. We think there are two reasonable explanations. On the one hand, the corner points can not provide sufficient supervision for the learning of the whole semantic response, and thus hamper the generating of pseudo boxes. On the other hand, in our method, the edge-aware neighbor cost is utilized to help stopping the diffusion of labeled points at boundaries, and thus prevent the labeled points from spreading towards the background areas (see Sec.~\ref{effect_ilg}). However, since the corner points are located on the boundaries of objects, the edge-aware cost may hinder the diffusion of the labeled points to the internal area of the object, as their paths pass through the edges. For example, as shown in the $P_{map}$ of the instance 6 in Fig.~\ref{fig_ilg}, the ILG module can recognize its correct regions with the semantic-aware cost only. However, when the edge-aware cost is introduced, the labeled points can only be diffused to background areas.}

\begin{table}[t]
	\begin{center}
		{
			\caption{Comparison of the pseudo box quality and detection performance achieved by our PLUG with different point label selection strategies.}
			%     \vspace{-0.2cm}
			\label{tbl_pointselect}
			%		\scriptsize
			\renewcommand\arraystretch{1.3}
			\setlength{\tabcolsep}{1.3mm}{
				\begin{tabular}{ccccc cccc}
					\hline
					Selection&\multicolumn{4}{c}{Pseudo box quality}&\multicolumn{4}{c}{Detection performance}\\
					\cline{2-9}
					Strategy&$\textit{mIoU}$ & $\textit{mIoU}_\textit{s} $ &$\textit{mIoU}_\textit{m}$&$\textit{mIoU}_\textit{l}$ & \textit{mAP}$_\textit{50}$ & \textit{mAP}$_\textit{s}$ & \textit{mAP}$_\textit{m}$ & \textit{mAP}$_\textit{l}$ \\
					\hline
					corner&0.518&0.488&0.586&0.589&0.406&0.306&0.464&0.427\\
					center&0.553&0.520&0.629&0.606&0.438&0.316&0.493&0.504\\
					random&0.549&0.539&0.576&0.533&0.427 &0.329&0.474&0.338\\
					\hline
		\end{tabular}}}
		%   \vspace{-0.2cm}
	\end{center}
\end{table}

{\subsection{Extension to Rotated Object Detection}}

{
	In our method, the ILG module utilize semantic and edge information to assign pixels to its most likely object or background, and use the circumscribed rectangle of assigned pixels as pseudo boxes. 
	Therefore, by further transforming the circumscribed rectangle to the one with the minimum area, our method can be easily extended to the task of rotated object detection. we conduct experiments to validate the effectiveness of our method on rotated object detection. Specifically, we use the modified PLUG to generate rotated pseudo boxes, and use ROITrans \cite{ding2019learning} as the downstream rotated detector to develop PLUG-ROITrans. The experimental results of our PLUG-ROITrans (under single point supervision) and the original ROITrans (under ground-truth rotated box supervision) are shown in Table~\ref{tbl_rotate}. It can be observed that our PLUG-ROITrans can achieve 0.351 in terms of $\textit{mAP}_\textit{50}$, which is 51.6\% of the performance of fully supervised ROITrans. The results demonstrate the preliminary effectiveness of our method in pointly supervised rotated object detection in RS images.}
%In the future, we will further investigate the rotation issue in pointly supervised RSOD.

\begin{table*}[htp]
	\begin{center}
		{
		\caption{Comparison of the detection performance achieved by ROITrans and PLUG-ROITrans.}
		%     \vspace{-0.2cm}
		\label{tbl_rotate}
				\scriptsize
		\renewcommand\arraystretch{1.3}
		\setlength{\tabcolsep}{1.3mm}{
			\begin{tabular}{rcc ccccc ccccc ccccc cc}
				\hline
				\multirow{2}{*}{Method}&\multirow{2}{*}{Supervision}&\multirow{2}{*}{Backbone}&\multicolumn{15}{c}{Categories} & \multirow{2}{*}{$\textit{mAP}_\textit{50}$}\\
				\cline{4-18}
				& & & PL &BD& BR & GTF & SV& LV&SH &TC &BC & ST & SBF & RA & HB & SP & HC&\\
				\hline
				PLUG-ROITrans&Point&ResNet50&0.088&0.503&0.302&0.292&0.248&0.661&0.368&0.806&0.285&0.502&0.400&0.365&0.259&0.176&0.009&0.351\\
				ROITrans \cite{ding2019learning}&Rotated Box&ResNet50&0.798&0.671&0.500&0.736&0.713&0.851&0.885&0.906&0.551&0.693&0.620&0.651&0.676&0.578&0.366&0.680\\
				\hline
		\end{tabular}}}
		%   \vspace{-0.2cm}
	\end{center}
\end{table*}

\subsection{Further Analyses on Dense Objects}
\label{motivation}

\begin{figure}[t]  
	\centering
	\includegraphics[width= 8.5 cm]{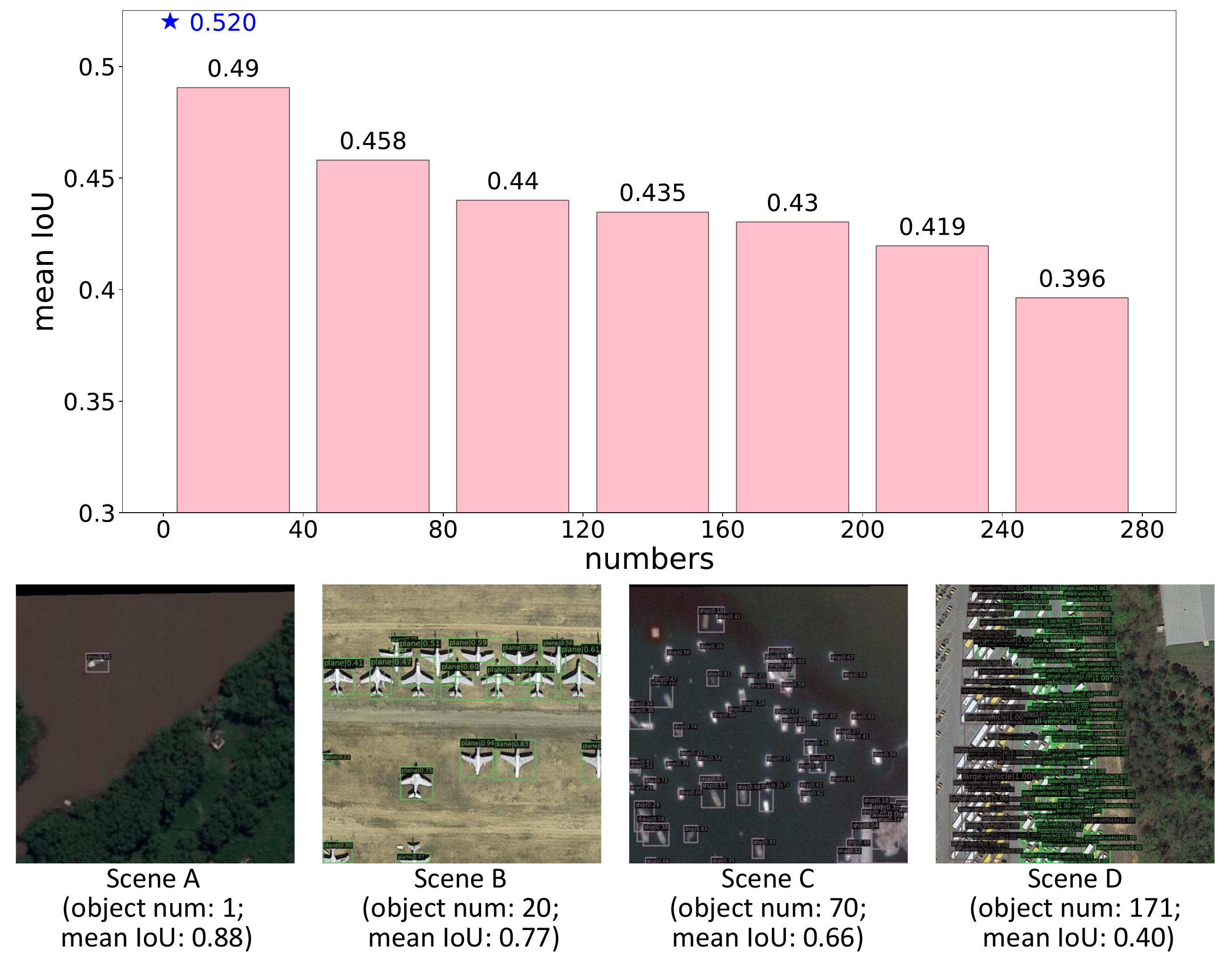}
	%  \vspace{-0.2cm}
	\caption{The $\textit{IoU}$ of generated pseudo boxes in images with different numbers. Here, four exampled scenes are shown for visualization. Note that, the blue star indicates that the mean $\textit{IoU}$ of pseudo boxes is 0.520 in images with single object.}
%		The $\textit{IoU}$ of generated pseudo boxes in images with different numbers. Note that, the results of our method w/o sparse feature guidance scheme are shown, and our method can improve the qualities of generated pseudo labels in different density intervals. Here, four exampled scenes are shown for visualization.
%		 Note that, the blue star indicates that the mean $\textit{IoU}$ of pseudo boxes is 0.520 in images with single object.}
	%  \vspace{-0.3cm}
	\label{fig_numiou}
	%	\vspace{-0.2cm}
\end{figure}

\begin{figure*}[t]
	\centering
	\includegraphics[width= 18 cm]{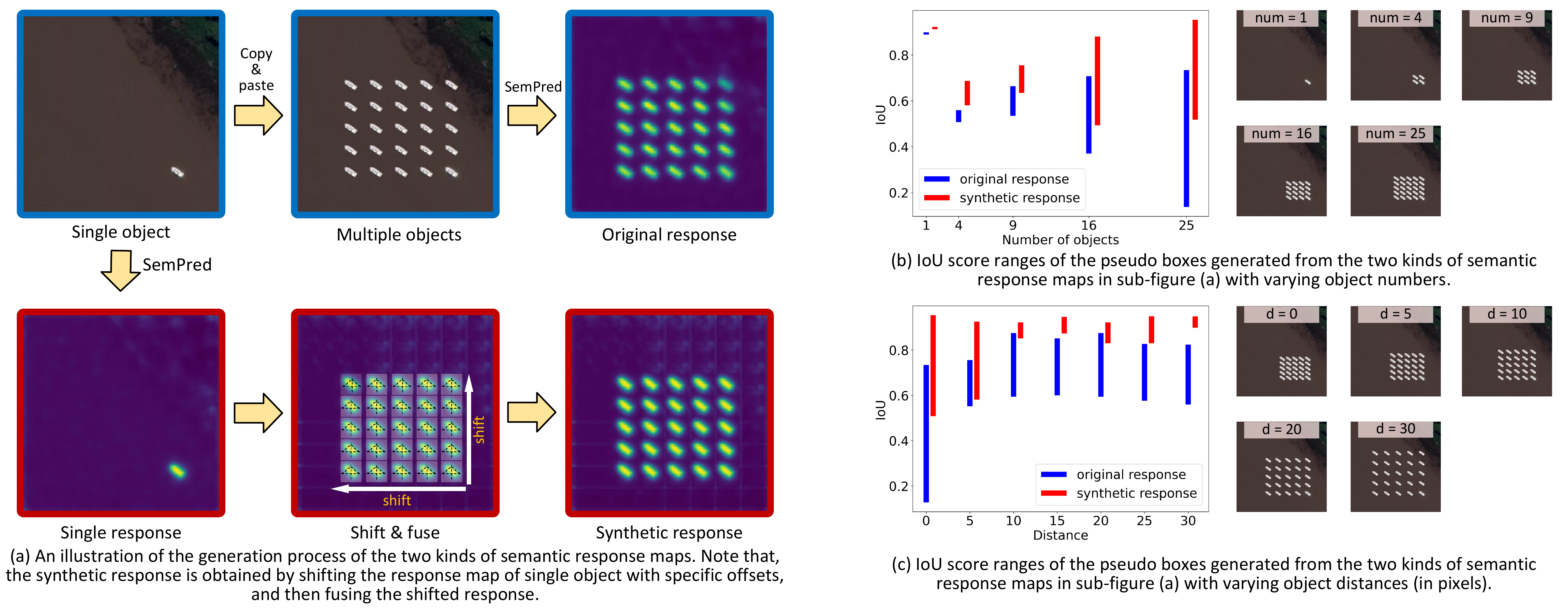}
	%  \vspace{-0.3cm}
	\caption{Illustrations and analyses of the influence of densely packed objects to the quality of generated pseudo boxes.}  
	%  \vspace{-0.2cm}
	\label{fig_feadisnumcopy}
	%	\vspace{-0.2cm}
\end{figure*}

As mentioned in Introduction, dense objects introduce challenges to discriminative feature extraction, and thus affect the quality of generated pseudo boxes. In this subsection, we conduct a series of experiments to analyze the influence of dense objects.

First, we coarsely suppose that the density of objects are positively related to their numbers in an image patch (with same area). Then, we split the DOTA dataset into several sub-sets containing different number of objects, and quantitatively evaluate the quality of generated pseudo boxes with respect to the object density. Note that, we do not perform sparse feature guidance in our PLUG to better demonstrate the challenges introduced by dense objects. As shown in Fig.~\ref{fig_numiou}, the quality of generated boxes degrades as the number of objects (i.e., density) increases. The examples in Scenes A to D qualitatively illustrate the quality degradation of pseudo boxes with dense objects.

Second, considering that the number, adjacent distance and appearance of objects are the three key factors that influence the quality of pseudo boxes, we design specific experiments to quantitatively investigate the impact of the first two factors by keeping the object appearance unchanged. Specifically, we use the ``copy-and-paste" strategy (see the sub-figures with blue boxes in Fig.~\ref{fig_feadisnumcopy}(a)) to generate multiple identical objects with controllable density. As shown in Fig.~\ref{fig_feadisnumcopy}(b) and \ref{fig_feadisnumcopy}(c), the quality of generated boxes degrades as the object density increases. 

Finally, we keep the density of the semantic response maps unchanged and investigate the influence of densely packed objects to the discriminative feature extraction. Specifically, we shift and fuse the single-object response to synthesize a pseudo dense-object response map. In this way, we build a control group with identical object density in the response maps but different feature representations in the feature extraction module. As shown in Fig.~\ref{fig_feadisnumcopy}(b) and Fig.~\ref{fig_feadisnumcopy}(c), the \textit{mIoU} scores of the pseudo boxes generated from the control group are significantly higher than those obtained from the images with dense objects. The experimental results clearly validate that densely packed objects in RS images can hinder the discriminative feature extraction and thus degrade the quality of pseudo boxes. {With our sparse feature guidance scheme, the $\textit{mIoUs}$ of generated pseudo labels in different density intervals are increased. The qualitative results demonstrate the effectiveness of our method in handling the densely packed objects.}

\begin{table*}
	\begin{center}
		\caption{Quantitative results achieved by different instance segmentation methods on the DOTA dataset.}
%   \vspace{-0.2cm}
		\label{tbl_psis_val}
		\renewcommand\arraystretch{1.3}
		\setlength{\tabcolsep}{3mm}{
			\begin{tabular}{r cccc c cccc}
				\hline
				\multirow{2}{*}{model} &\multirow{2}{*}{supervision}&\multicolumn{4}{c}{object detection}&\multicolumn{4}{c}{instance segmentation}\\
				\cline{3-10}
				& &\textit{mAP}$_\textit{50}$ &\textit{mAP}$_\textit{s}$ &\textit{mAP}$_\textit{m}$&\textit{mAP}$_\textit{l}$&\textit{mAP}$_\textit{50}$ &\textit{mAP}$_\textit{s}$ &\textit{mAP}$_\textit{m}$&\textit{mAP}$_\textit{l}$\\
				\hline			
				
				\rowcolor{gray!5}
    Mask-RCNN \cite{MaskRCNN}&\textit{Mask}&0.659&0.535&0.670&0.697&0.623&0.480&0.662&0.682 \\
               \rowcolor{gray!5} 
               BoxInst \cite{Boxinst} &\textit{Box}&0.643 &0.535&0.633&0.647&0.503&0.371&0.543&0.582 \\
                \rowcolor{green!15}
                PLUG-Seg (ours)&\textit{Point}&0.435&0.335&0.481&0.348&0.406&0.278&0.491&0.340\\
				\hline
		\end{tabular}}
	\end{center}
%  \vspace{-0.2cm}
\end{table*}
\begin{figure*}[t]
	\centering
	\includegraphics[width=18cm]{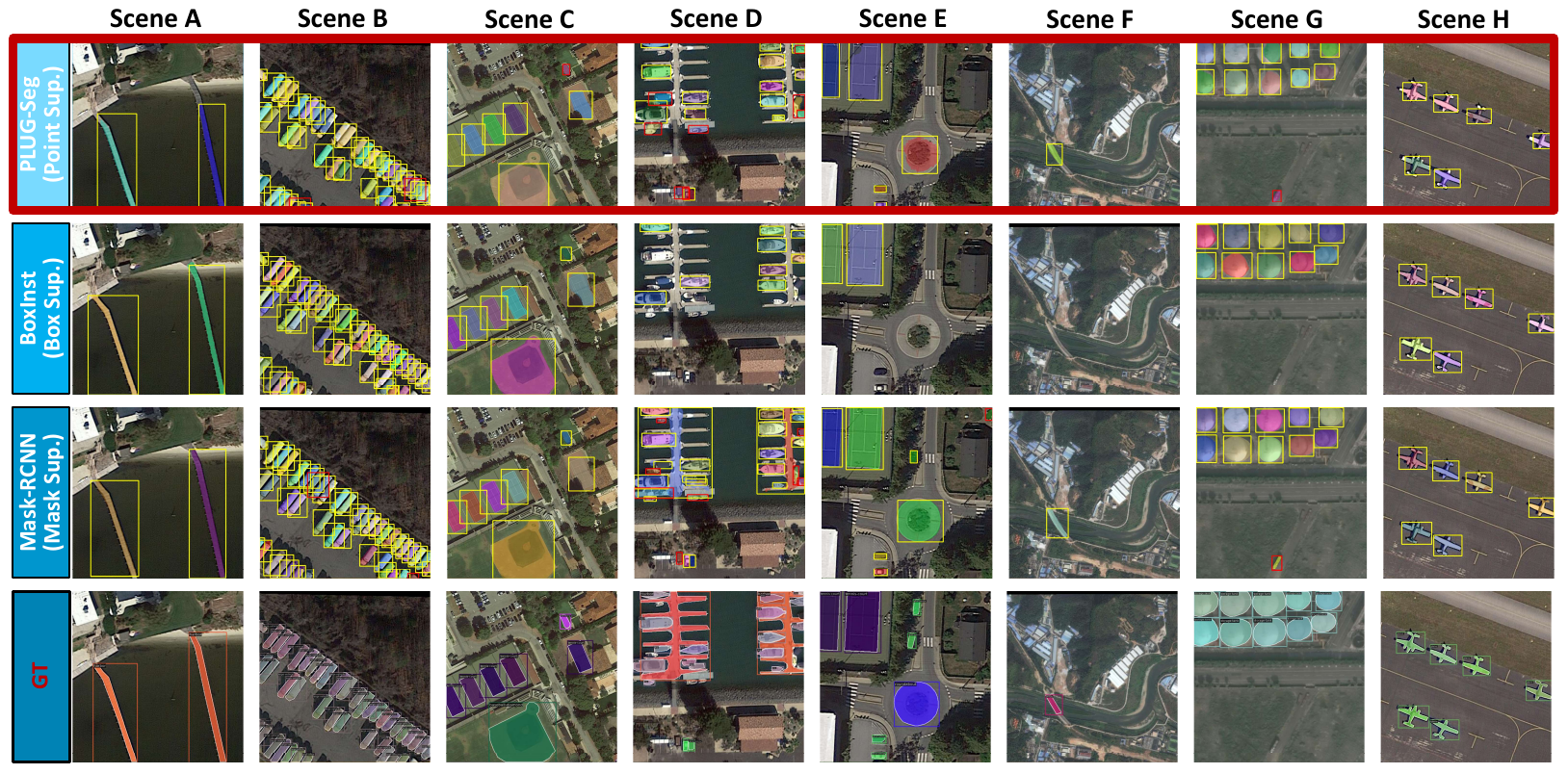}
%  \vspace{-0.2cm}
	\caption{Qualitative results achieved by different instance segmentation methods on the DOTA dataset. The correctly detected results are marked by yellow boxes, and the falsely detected results are marked by red boxes. Gradually darker colors represent stronger supervision. The predicted instance masks are randomly colored.}
%  \vspace{-0.2cm}
	\label{fig_segvisual}
	%	\vspace{-0.2cm}
\end{figure*}

\subsection{Extension to Instance Segmentation}

Since the ILG module in our PLUG produces instance label for each object from their point annotation, our PLUG can be easily extended to pointly supervised instance segmentation (PSIS). Specifically, we concatenate our PLUG with Mask-RCNN, and developed a PLUG-Seg network to achieve PSIS in RS images. Besides, we used the groundtruth mask labels in the iSAID dataset \cite{waqas2019isaid}, and adopted the mask-level \textit{mAP}$_\textit{50}$, \textit{mAP}$_\textit{s}$, \textit{mAP}$_\textit{m}$ and \textit{mAP}$_\textit{l}$ as quantitative metrics for performance evaluation. We compare our PLUG-Seg with BoxInst \cite{Boxinst} and Mask-RCNN \cite{MaskRCNN}, which use box-level and mask-level supervision for instance segmentation, respectively. We also followed these two methods \cite{Boxinst,MaskRCNN} to evaluate the performance of object detection and instance segmentation simultaneously. The experimental results are shown in Table~\ref{tbl_psis_val} and Fig.~\ref{fig_segvisual}.

It can be observed from Table~\ref{tbl_psis_val} that our PLUG-Seg can achieve an \textit{mAP}$_\textit{50}$ of 0.435 for object detection and an \textit{mAP}$_\textit{50}$ of 0.406 for instance segmentation. With single point annotation for each instance, our PLUG-Seg can achieve 68\%$/$81\% and 66\%$/$65\% accuracy in object detection$/$instance segmentation as compared to box-level (i.e., BoxInst \cite{Boxinst}) and mask-level (i.e., Mask-RCNN \cite{MaskRCNN}) supervised methods, respectively. 
The qualitative results in Fig.~\ref{fig_segvisual} also demonstrate the promising performance of our PLUG-Seg. It is worth noting that our PLUG-Seg can achieve better performance than BoxInst \cite{Boxinst} on scenes with complex backgrounds (e.g., the roundabout and small vehicles in Scene E and the bridge in Scene F). These experimental results demonstrate that single point annotation can provide sufficient supervision for instance segmentation. 

\section{Conclusion}\label{sec:conclusion}
In this paper, we proposed a method to learn remote sensing object detection  with single point supervision. In our method, a point label upgrader (PLUG) is designed to generate pseudo boxes from point labels. We also handle the dense object issue in remote sensing images by designing a sparse feature guided semantic prediction module. Experimental results validate the effectiveness and superiority of our method. \textcolor{black}{In the future, we will further extend our method to generate rotated pseudo boxes from single point labels, and investigate more stable and efficient pseudo label generation schemes.}
We hope our study can draw attention to the research of single pointly supervised remote sensing object detection.
%\cite{chen2021points, zhang2022group, zhang2023weakly, ge2023point, li2023monte, ying2023mapping, bearman2016s}
%\cite{qian2019weakly, wu2022deep, laradji2020proposal, cheng2022pointly, liao2023attentionshift, fan2022pointly}
%\cite{ribera2019locating, song2021rethinking, yu2022object}

\bibliographystyle{IEEEtran}
\bibliography{PLUG-Det}

\begin{IEEEbiography}	[{\includegraphics[width=1in,height=1.25in,clip,keepaspectratio]{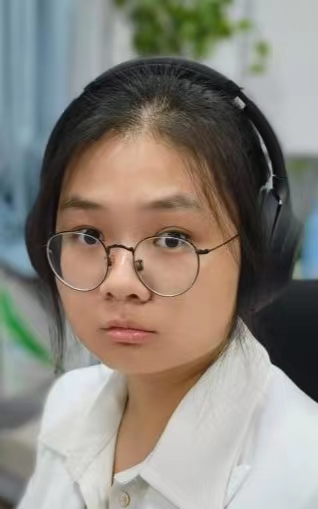}}]{Shitian He} received the B.E. degree in electronic and information engineering from Beijing Jiaotong University (BJTU), Beijing, China, in 2019, and the M.E. degree in information and communication engineering from National University of Defense Technology (NUDT), Changsha, China, in 2021. She is currently pursuing the Ph.D. degree with the College of Electronic Science and Technology, NUDT. Her research interests focus on object detection and tracking.	
\end{IEEEbiography}
\vspace{-1cm}

\begin{IEEEbiography}
	[{\includegraphics[width=1in,height=1.25in,clip,keepaspectratio]{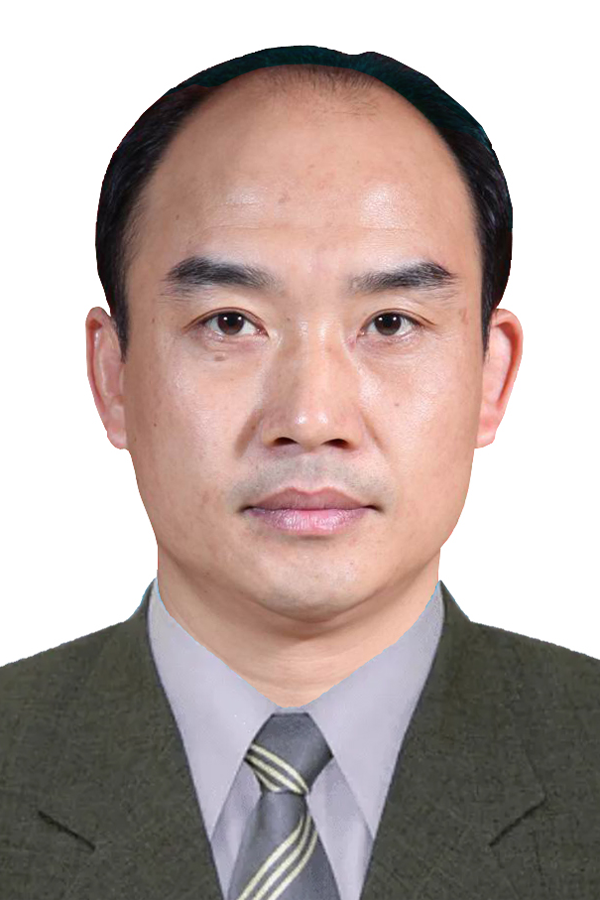}}]{Huanxin Zou} was born in Meizhou, Guangdong, China in 1973. He received the B.S. degree in information engineering from the National University of Defense Technology, Changsha, Hunan, China in 1995, and the M.S. and Ph.D. degree in information and communication engineering from the National University of Defense Technology, Changsha, Hunan, China in 2000 and 2003, respectively. He is currently a professor with the College of Electronic Science and Technology, National University of Defense Technology, China. He held a visiting position with the Department of Computing Science in the University of Alberta, Canada, from March to September in 2015. His main research interests include computer vision, deep learning, pattern recognition, and remote sensing image processing and interpretation.
\end{IEEEbiography}
\vspace{-1cm}

\begin{IEEEbiography}
	[{\includegraphics[width=1in,height=1.25in,clip,keepaspectratio]{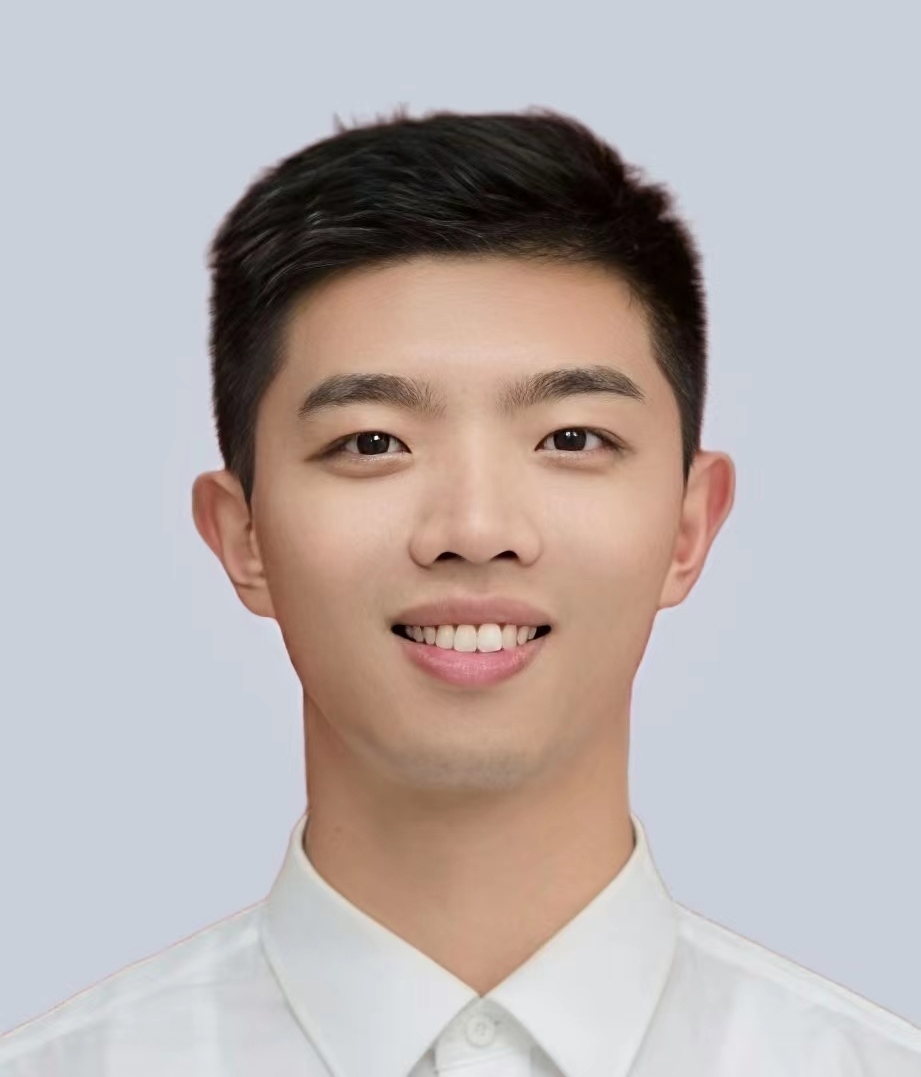}}]{Yingqian Wang} received the B.E. degree in electrical engineering from Shandong University (SDU), Jinan, China, in 2016, the M.E. degree and the D.E. degree in information and communication engineering from National University of Defense Technology (NUDT), Changsha, China, in 2018 and 2023, respectively. He is currently a lecturer with the College of Electronic Science and Technology, NUDT. His research interests focus on low-level vision, particularly on light field imaging and image super-resolution.
\end{IEEEbiography}
\vspace{-1cm}

\begin{IEEEbiography}
	[{\includegraphics[width=1in,height=1.25in,clip,keepaspectratio]{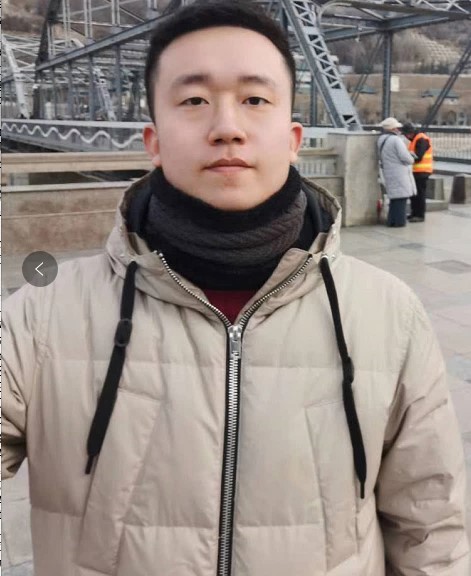}}]{Boyang Li} received the B.E. degree in Mechanical Design manufacture and Automation from the Tianjin University, China, in 2017 and M.S. degree in biomedical engineering from National Innovation Institute of Defense Technology, Academy of Military Sciences, Beijing, China, in 2020. He is currently working toward the PhD degree in information and communication engineering from National University of Defense Technology (NUDT), Changsha, China. His research interests include infrared small target detection, weakly supervised semantic segmentation and deep learning.
\end{IEEEbiography}
\vspace{-1cm}

\begin{IEEEbiography}	[{\includegraphics[width=1in,height=1.25in,clip,keepaspectratio]{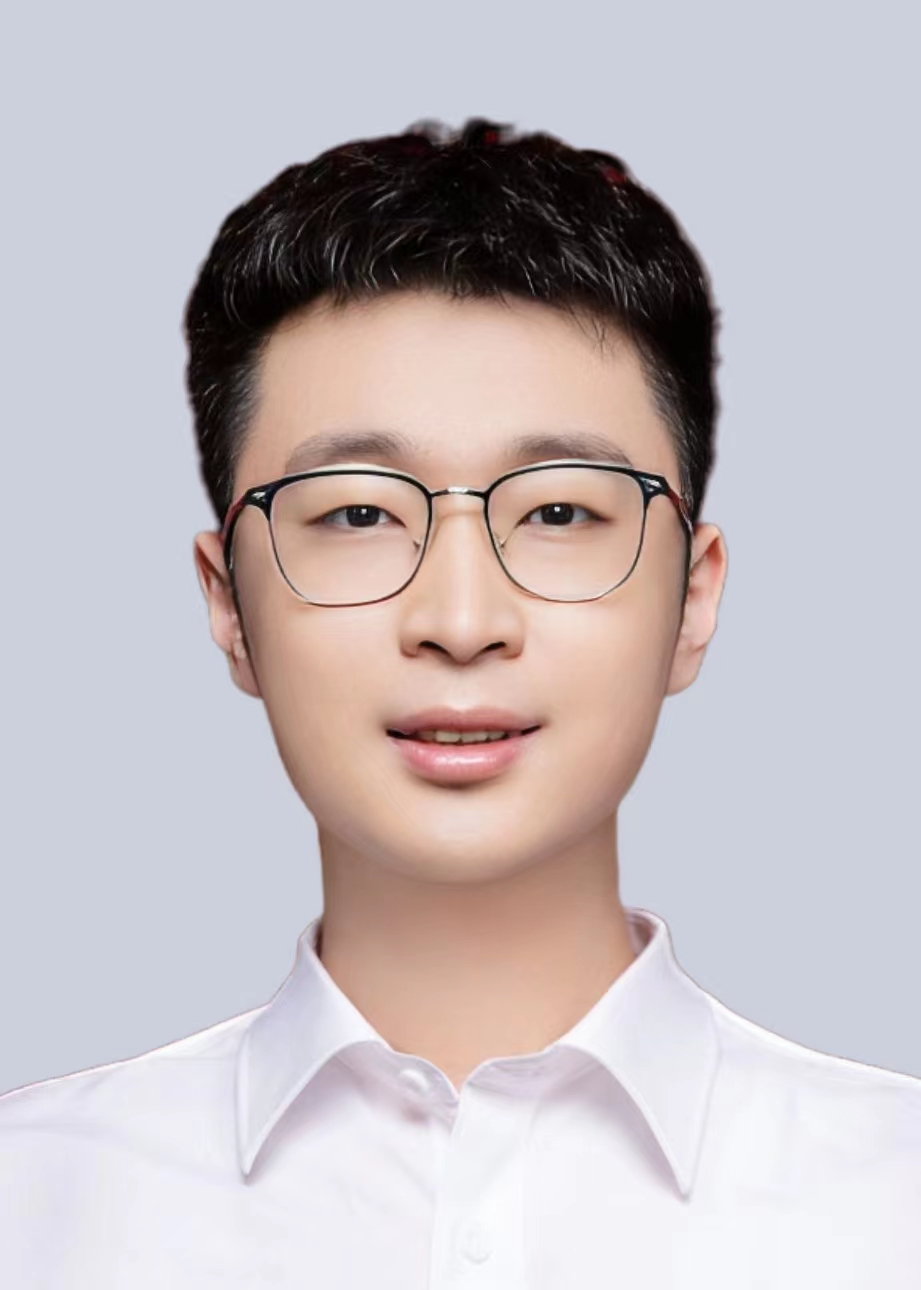}}]{Xu Cao} received the B.E. degree in information engineering in 2018, and the M.E. degree in information and communication engineering from National University of Defense Technology (NUDT), Changsha, China, in 2021. He is currently pursuing the Ph.D. degree with the College of Electronic Science and Technology, NUDT. His research interests focus on object detection and multi-source information fusion interpretation.
\end{IEEEbiography}
\vspace{-1cm}
% \begin{IEEEbiography}
% 	[{\includegraphics[width=1in,height=1.25in,clip,keepaspectratio]{Bio/NingJIng.jpg}}]{Ning Jing} received the BSc degree in communication and information systems, the MSc degree in signal and information processing, and the PhD degree in computer science from the National University of Defense Technology, China, in 1983, 1986, and 1990, respectively. He is currently a professor at the National University of Defense Technology, China. His research interests include GIS, database systems, spatial data analysis, and visualization. He is a senior fellow of the China Computer Federation (CCF), a fellow of the Technical Committee of Database System of CCF and a fellow of the Technical Committee of Principles and Methods of the China GIS Association.
% \end{IEEEbiography}
\begin{IEEEbiography}
	[{\includegraphics[width=1in,height=1.25in,clip,keepaspectratio]{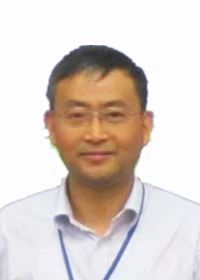}}]{Ning Jing} received the PhD degree in Computer Science from National University of Defense Technology, in 1990. He is currently a professor in National University of Defense Technology. His research interests include management and analysis of big data, spatial information system.
\end{IEEEbiography}

\end{document}